\documentclass[10pt,twocolumn,letterpaper]{article}

\usepackage{cvpr}              

\usepackage{mathtools}
\usepackage{enumitem}
\usepackage[accsupp]{axessibility}  
\usepackage{graphicx}
\usepackage{amsmath}
\usepackage{amssymb}
\usepackage{booktabs}
\usepackage{esvect}
\usepackage{pifont}
\usepackage[pagebackref,breaklinks,colorlinks]{hyperref}

\usepackage[capitalize]{cleveref}
\crefname{section}{Sec.}{Secs.}
\Crefname{section}{Section}{Sections}
\Crefname{table}{Table}{Tables}
\crefname{table}{Tab.}{Tabs.}

\def\cvprPaperID{1022} 
\def\confName{CVPR}
\def\confYear{2023}

\begin{document}

\title{Learning Locally Editable Virtual Humans}

\author{Hsuan-I Ho
\and
Lixin Xue
\and
Jie Song
\and
Otmar Hilliges
\and
Department of Computer Science, ETH Zürich}

\maketitle

\begin{abstract}
In this paper, we propose a novel hybrid representation and end-to-end trainable network architecture to model fully editable and customizable neural avatars. 
At the core of our work lies a representation that combines the modeling power of neural fields with the ease of use and inherent 3D consistency of skinned meshes. 
To this end, we construct a trainable feature codebook to store local geometry and texture features on the vertices of a deformable body model, thus exploiting its consistent topology under articulation. 
This representation is then employed in a generative auto-decoder architecture that admits fitting to unseen scans and sampling of realistic avatars with varied appearances and geometries. 
Furthermore, our representation allows local editing by swapping local features between 3D assets.
To verify our method for avatar creation and editing, we contribute a new high-quality dataset, dubbed CustomHumans, for training and evaluation. Our experiments quantitatively and qualitatively show that our method generates diverse detailed avatars and achieves better model fitting performance compared to state-of-the-art methods.
Our code and dataset are available at \url{https://custom-humans.github.io/}. 
\end{abstract}
\vspace{-1em}


\begin{figure}[t]
\centering
\captionsetup{type=figure}
\includegraphics[width=\linewidth]{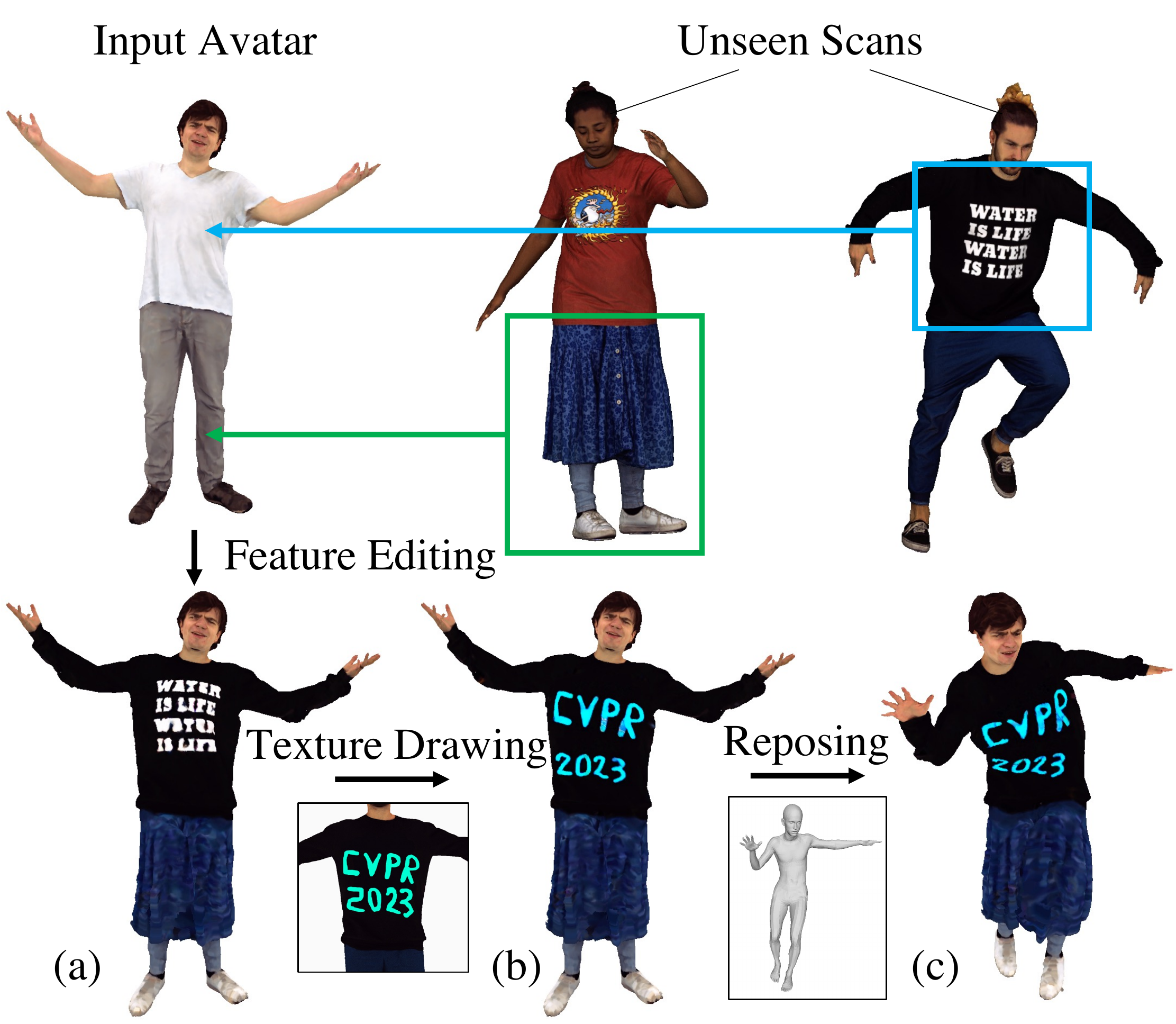}
\caption{\textbf{Creating locally editable avatars:} Given an input avatar, (a) the avatar can be edited by transferring clothing geometry and color details from existing, yet unseen 3D assets. (b) Users can customize clothing details such as logos and letters via drawing on 2D images. (c) The avatars retain local detail consistently under pose changes.}
\label{fig:teaser}
\vspace{-2em}
\end{figure}







\section{Introduction}
\label{sec:intro}
3D Avatars are an important aspect of many emerging applications such as 3D games or the Metaverse. 
Allowing for easy personalization of such avatars, holds the promise of increased user engagement.
Traditionally, editing 3D assets requires knowledge of computer graphics tools and relies on standardized data formats to represent shapes and appearances. 
While methods for reconstruction or generative modeling of \emph{learned} avatars achieve impressive results, it is unknown how such neural avatars can be edited and customized. 
Thus, the goal of our work is to contribute a simple, yet powerful data-driven method for avatar creation and customization~(\cref{fig:teaser}):
our method enables
(a) the ability to transfer partial geometric and appearance details between 3D assets, 
and (b) the ability to author details via 2D-3D transfer. 
The resulting avatars (c) retain consistent local details when posed.




Existing methods do not allow for such capabilities. 
While 3D generative models of articulated human bodies~\cite{EVA3D,bergman2022generative, zhang2022avatargen, noguchi2022unsupervised, grigorev2021stylepeople} leverage differentiable neural rendering to learn from images, they cannot control local details due to highly entangled color and geometry in the 2D supervision signal. 
Generative models trained on 3D data~\cite{CAPE:CVPR:20, corona2021smplicit, palafox2021npms, palafox2021spams, chen2022gdna} can produce geometric details for surfaces and clothing. However, the diversity of generated samples is low due to the lack of high-quality 3D human scans and not all methods model appearance.


At the core of the issue lies the question of representation: graphics tools use meshes, UV, and texture maps which provide consistent topologies under deformation. However, human avatar methods that are built on mesh-based representations and linear blend skinning (LBS) are limited in their representational power with respect to challenging geometry (e.g., puffy garments) and flexible topologies (e.g., jackets), even with adaptations of additional displacement parameters~\cite{CAPE:CVPR:20} and mesh subdivision~\cite{tiwari20sizer}.  



Inspired by the recent neural 3D representations~\cite{takikawa2021nglod,mueller2022instant,yang2022neumesh,xie2022neural}, we propose a novel \emph{hybrid} representation for digital humans. 
Our representation combines the advantages of consistent topologies of LBS models with the representational power of neural fields. 
The key idea is to decompose the tasks of deformation consistency on one hand and local surface and appearance description on the other. 
For the former, we leverage existing parametric body models (e.g., SMPL~\cite{SMPL:2015} and SMPL-X~\cite{pavlakos2019expressive}). 
For the latter, we leverage the fixed topology of the poseable mesh to store local feature codebooks. 
A decoder, shared across subjects, is then conditioned on the local features to predict the final signed distance and color values. 
Since only local information~\cite{devries2021unconstrained} is exposed to the decoder, overfitting and memorization can be mitigated.  
We experimentally show that this is crucial for 3D avatar fitting and reposing. 

Complementing this hybrid representation, we propose a training pipeline in the auto-decoding generative framework~\cite{chen2022gdna,park2019deepsdf,rebain2022lolnerf}. 
To this end, we jointly optimize multi-subject feature codebooks and the shared decoder weights via 3D reconstruction  and 2D adversarial losses. 
The 3D losses help in disentangling appearance and geometric information from the input scans, while the latter improves the perceptual quality of randomly generated samples. 
To showcase the hybrid representation and the generative model we implement a prototypical avatar editing workflow shown in Fig.~\ref{fig:teaser}.

Furthermore, to enable research on high-quality 3D avatars we contribute training data for generative 3D human models. We record a large-scale dataset (more than 600 scans of 80 subjects in 120 garments) using a volumetric capture stage~\cite{collet2015high}. Our dataset consists of high-quality 3D meshes alongside accurately registered SMPL-X~\cite{pavlakos2019expressive} models and will be made available for research purposes. Finally, we assess our design decisions in detailed evaluations, both on existing and the proposed datasets.


In summary, our contributions are threefold: (a) a novel hybrid representation for 3D virtual humans that allows for local editing across subjects, (b) a generative pipeline of 3D avatars creation that allows for fitting to unseen 3D scans and random sampling, and (c) a new large-scale high-quality dataset of 3D human scans containing diverse subjects, body poses and garments. 

\begin{figure*}[t]
\centering
\includegraphics[width=\linewidth]{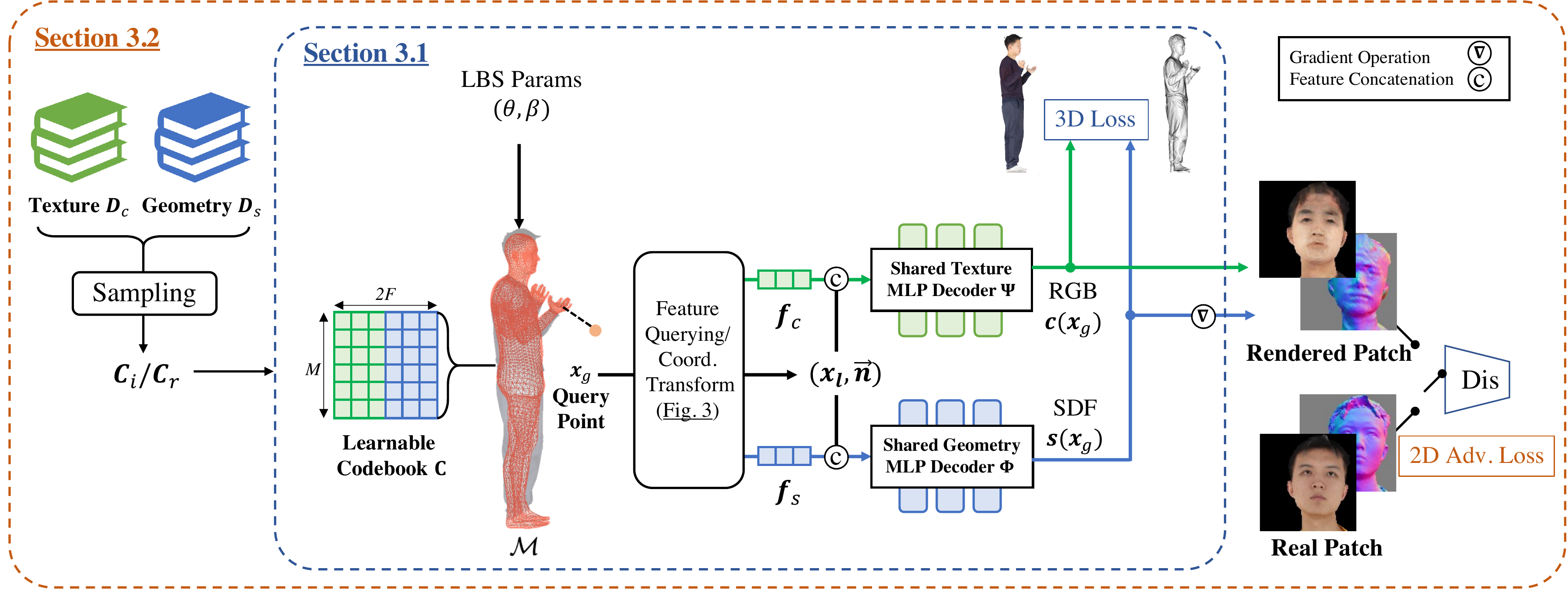}
\caption{\textbf{Proposed framework}. 
Given a posed scan registered with body pose and shape LBS parameters $(\theta,\beta)$, our proposed human representation stores its local geometry and texture features in a codebook $\mathbf{C}$ which is indexed by the vertex indices of $M$-vertex LBS body mesh  $\mathcal{M}$ (\cref{sec:representation}). Given a query point coordinates $\mathbf{x}_g$, two separated MLP decoders $\mathbf{\Psi, \Phi}$ predict signed distances and colors conditioned on the positional features ($\mathbf{x}_l, \vec{\mathbf{n}}$) and the local geometry/texture features ($\mathbf{f_s}$ / $\mathbf{f_c}$) respectively. We train a generative auto-decoder using $N$ posed scans, whose feature codebooks are stored in the dictionary $\mathbf{D_s}, \mathbf{D_c}$. We introduce two sampling strategies to sample codebooks (denoted as $\mathbf{C}_i$/$\mathbf{C}_r$ respectively) and jointly train our dictionaries and shared MLP decoders with a 3D reconstruction loss and a 2D adversarial loss (\cref{sec:sampling} \&~\cref{sec:training}). }
\label{fig:framework}
\vspace{-1.5em}
\end{figure*}
\section{Related Work}
\label{sec:work}

\paragraph{Controllable human representations.}
Topics of virtual humans have received much attention in the graphics literature, such as skinning and rigging of articulated meshes~\cite{SMPL:2015,MANO:SIGGRAPHASIA:2017, pavlakos2019expressive,osman2020star}, physical simulation of clothing~\cite{goldenthal2007efficient,narain2012adaptive,grigorev2022hood}, and deferred rendering~\cite{thies2019deferred, pedersen1995decorating, hanrahan1990direct}. With the advances in neural rendering~\cite{tewari2020state, tewari2022advances} and the availability of large-scale human datasets~\cite{lassner2017generative,liu2016deepfashion,fu2022stylegan,li2021learn,patel2021agora,varol2017learning,yu2020humbi,tao2021function4d}, numerous approaches have been proposed to reconstruct~\cite{saito2020pifuhd, saito2019pifu, alldieck2022phorhum} and explicitly control~\cite{liu2021neural, peng2021neural, chen2021snarf} human avatars in a data-driven manner.

One branch of work focuses on 2D image synthesis via generative adversarial networks (GANs)~\cite{goodfellow2014generative} and techniques of feature manipulation~\cite{lassner2017generative,shi2021lifting,roich2022pti}. Typically, a 2D neural renderer creates human images corresponding to pose and appearance latent codes learned from the training data. 
Related applications such as virtual try-on~\cite{xie2021towards, dong2019fw, Dong_2022_CVPR} and video retargeting~\cite{wang2021one, chan2019everybody, liu2019liquid} have shown promising results in light of photo-realistic image synthesis by GANs~\cite{fu2022stylegan, karras2021alias}. However, these methods do not explicitly reason about complex 3D human geometry and can therefore not produce 3D-consistent results. 

A newly emerging line of work aims to create controllable avatars with 3D consistency. Some methods extend existing body models with neural networks to predict displacement layers~\cite{alldieck2019learning, bhatnagar2020combining,CAPE:CVPR:20} or textures~\cite{prokudin2021smplpix}.
Other methods learn to model challenging pose-dependent deformations on avatars either by predicting LBS weights~\cite{deng2019neural, chen2021snarf, saito2021scanimate, zheng2022imavatar, chen2022fastsnarf, shen2023xavatar} or improving the capabilities of body models~\cite{remelli2022drivable, peng2021neural, liu2021neural, POP:ICCV:2021, Ma:CVPR:2021, SkiRT:3DV:2022, guo2023vid2avatar} with the power of implicit neural fields. However, these approaches mainly focus on modeling a \emph{single} subject in specific clothing and do not scale to create diverse avatars. 
Our method overcomes this issue by learning a \emph{multi-subject} generative model which produces realistic virtual humans with disentangled controllability over body poses, clothing geometry, and texture.





\vspace{-1em}
\paragraph{Generative 3D human models.}
Existing generative models of human avatars can be loosely split into two main streams: learning 3D-aware neural rendering from 2D images~\cite{bergman2022generative, EVA3D, zhang2022avatargen, noguchi2022unsupervised, grigorev2021stylepeople} and learning body shapes from 3D supervision~\cite{CAPE:CVPR:20, corona2021smplicit, palafox2021npms, palafox2021spams, chen2022gdna}. Powered by recent advances in differentiable neural rendering~\cite{tewari2022advances} and neural fields~\cite{xie2022neural}, much progress has been made in 3D-aware generative models~\cite{shi2022deep}. However, learning to generate detailed clothed avatars from pure 2D supervision~\cite{EVA3D,bergman2022generative, zhang2022avatargen, noguchi2022unsupervised, grigorev2021stylepeople} is still challenging due to the complex appearance and articulation of bodies, self-occlusions, and highly entangled colors and geometries in images. 

More closely related to our setting are methods that learn to generate detailed body shapes from 3D scans or RGB-D data. For instance, CAPE~\cite{CAPE:CVPR:20} and SMPLicit\cite{corona2021smplicit} are generative models for clothed humans. The former exploits VAE-GAN to predict additive displacements based on the SMPL vertices while the latter drape an implicitly modeled garment layer onto SMPL. NPMs~\cite{palafox2021npms}, and SPAM~\cite{palafox2021spams} learn pose and shape latent spaces from 3D supervision, which enables latent code inversion using point clouds or depth sequences. gDNA~\cite{chen2022gdna} learns to synthesize body shapes in the canonical space and further improves clothing details with adversarial losses. However, none of the above-mentioned works is able to generate human bodies with appearance and neither allows fine-grained editing of the generated avatars. Our method addresses both issues by learning disentangled local representations for multiple subjects. In addition, we experimentally show that our representation significantly improves the performance of model fitting against state-of-the-art human generative models.


\section{Method}
Our proposed method is summarized in~\cref{fig:framework}. We first contribute a novel hybrid human representation that stores local geometric and textural information into two aligned feature spaces (\cref{sec:representation} and~\cref{fig:representation}). 
To allow fitting to new 3D scans and drawing random samples from the underlying data distribution, we design a training strategy to learn a meaningful latent space under the generative adversarial framework to bring in additional 2D adversarial supervision
(\cref{sec:sampling} \&~\cref{sec:training}).
Finally, we demonstrate the utility of our method for creating avatars by enabling local feature editing through existing 3D assets or images (\cref{sec:cutomize}). 




\subsection{Hybrid Representation of Humans}
\label{sec:representation}
To enable 3D avatars with high-fidelity representational power and local editing capabilities, a suitable representation is needed. 
To this end, we propose a novel \emph{hybrid} representation that combines the advantages of neural fields (flexibility and modeling power) with LBS-articulated mesh models (ease of deformation and full explicit control). 



An overview of how we leverage the proposed representation is provided in \cref{fig:framework} in the dotted blue box. 
Given a human scan, we first create a posed, coarse body mesh $\mathcal{M}$ (shown in red) using the registered body parameters $(\theta, \beta)$ of an LBS body model. 
The mesh consists of $M$ vertices ($\mathcal{V} \in \mathbb{R}^{M \times 3}$) in the posed space and $M_f$ faces where $\mathcal{F} \in \{ 1,...,M\}^{M_f\times 3}$ defines the vertex indices on each face. 
We then construct a trainable feature codebook $\mathbf{C}\in \mathbb{R}^{M\times2F}$, which stores $F$-dimensional local geometry and texture features respectively for each vertex.
%

\begin{figure}[t]
\centering
\includegraphics[width=\linewidth]{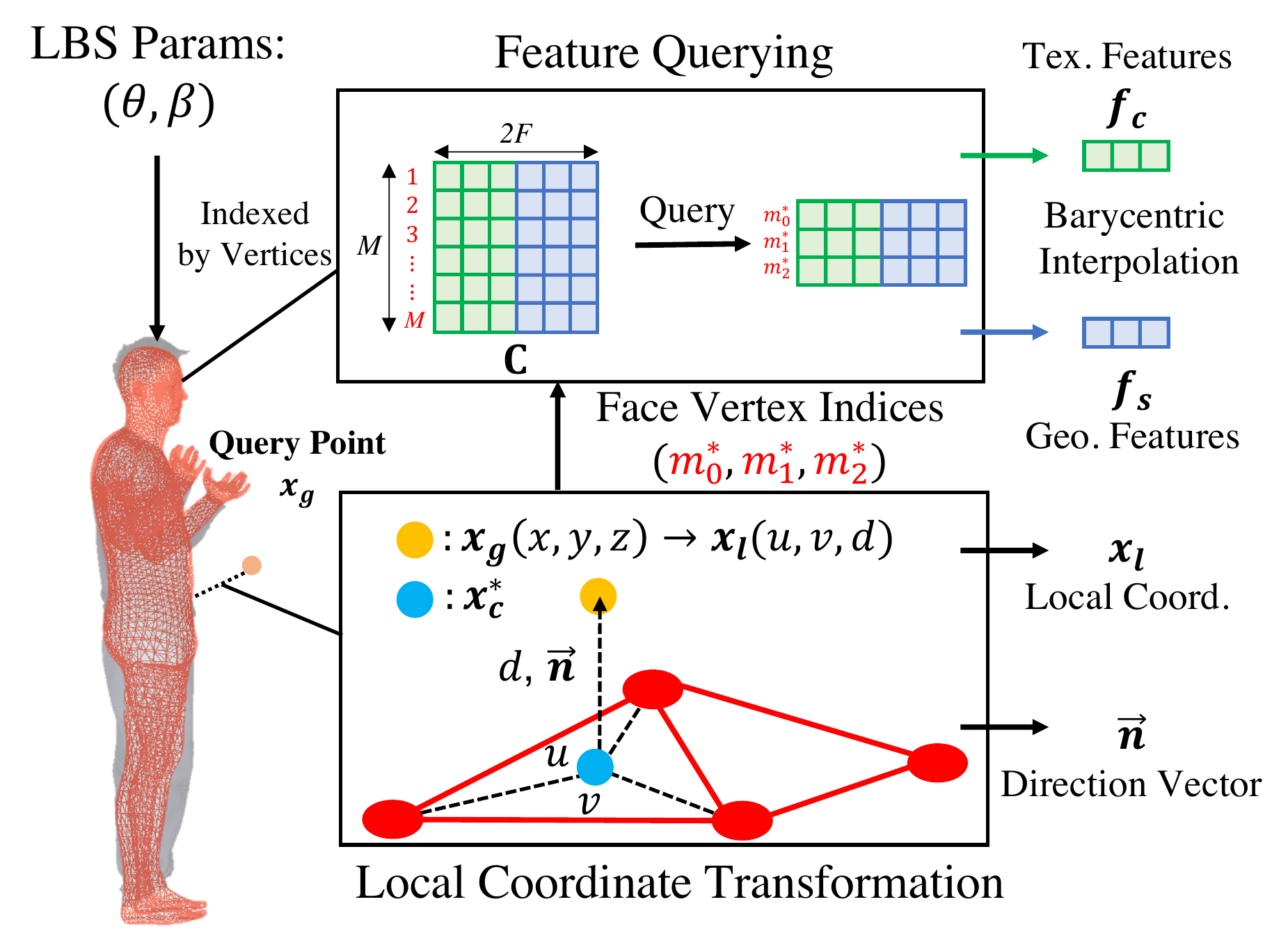}

\caption{\textbf{Local feature querying.} Given an LBS body mesh posed by the parameters ($\theta, \beta$), we represent a detailed human body as a codebook that stores local texture and geometry features indexed by the vertices on the mesh.
An input query point $\mathbf{x_g}$ finds the nearest triangle on the LBS body mesh and returns its vertex indices for local feature lookup. To prevent decoders from memorizing any global information, we transform the position of $\mathbf{x_g}$ into local triangle coordinates $(u, v)$, distance $d$, and direction $\vec{\mathbf{n}}$ of the closest point (i.e., the blue dot). The final geometry and texture features $\mathbf{f_s}, \mathbf{f_c}$ are fused via barycentric interpolation.
}
\label{fig:representation}
\vspace{-1.5em}
\end{figure}

Similar to coordinate-based implicit fields, a 3D coordinate $\mathbf{x}_g \in \mathbb{R}^{3}$ is used to predict its corresponding signed distance $s(\mathbf{x}_g)$, and RGB color $c(\mathbf{x}_g)$. 
Instead of using global coordinates directly as inputs, we condition neural field decoders on the local triangle coordinates $\mathbf{x}_l \in \mathbb{R}^{3}$ and the local geometry and texture features $\mathbf{f_s}, \mathbf{f_c} \in \mathbb{R}^F$. We illustrate this conversion from global coordinates to local triangle coordinates in \cref{fig:representation}. The global coordinates $\mathbf{x}_g$ are first projected onto the mesh by finding the closest point $\mathbf{x}_c^*$ (\cref{fig:representation}, blue dot):
\begin{equation}
\begin{aligned}
\mathbf{x}_c^* &= \arg\min_{\mathbf{x}_c}  \Vert \mathbf{x}_g - \mathbf{x}_c \Vert_{2}, \\
\mathbf{x}_c &= \mathcal{B}_{u,v} (\mathcal{V}[m_0, m_1, m_2]),
\end{aligned}
\end{equation}
where $(m_0, m_1, m_2)$ are vertex indices of faces $\mathcal{F}$ and $\mathcal{B}_{u,v}(.)$ is the barycentric interpolation function with barycentric coordinates $(u,v, 1-u-v)$.
The closest point $\mathbf{x}_c^*$ within the face is used to transform $\mathbf{x}_g$ into a local triangle coordinate system. Hence, $\mathbf{x}_l$ consists of the barycentric coordinates $(u, v)$, the signed distance $d$ between $\mathbf{x}_g$ and $\mathbf{x}_c^*$, i.e., $\mathbf{x}_l \coloneqq (u, v, d)$. We also compute a direction vector $\vec{\mathbf{n}}$ between $\mathbf{x}_g$ and $\mathbf{x}_c^*$ as an additional feature to distinguish points near triangle edges. 
To query local features $(\mathbf{f_s}, \mathbf{f_c})$, we use the vertex indices on the triangle $(m_0^*, m_1^*, m_2^*)$ to look up three elements in the feature codebook $\mathbf{C}$. We then fuse the three local features via barycentric interpolation.

Finally, we take all the local features $(\mathbf{f_s}, \mathbf{f_c}, \mathbf{x}_l, \vec{\mathbf{n}})$ as input to two separate decoders $\mathbf{\Phi}$ and $\mathbf{\Psi}$ to predict SDF and RGB values respectively:
\begin{equation}
\begin{aligned}
      \mathbf{\Phi} : \mathbb{R}^{F} \times \mathbb{R}^{3} \times \mathbb{R}^{3} &\rightarrow \mathbb{R} \\
     (\mathbf{f_s}, \mathbf{x}_l, \vec{\mathbf{n}}) &\mapsto s(\mathbf{x}_g), \\
\end{aligned}
\end{equation}
\begin{equation}
\begin{aligned}
      \mathbf{\Psi} : \mathbb{R}^{F} \times \mathbb{R}^{3} \times \mathbb{R}^{3} &\rightarrow \mathbb{R}^3 \\
     (\mathbf{f_c}, \mathbf{x}_l, \vec{\mathbf{n}}) &\mapsto c(\mathbf{x}_g). \\
\end{aligned}
\end{equation}
Note that only local information is exposed to the decoders, which allows us to use the same MLPs across different vertices and subjects. We show that preventing networks from memorizing global information in this way is necessary for local editing and reposing in our experiments (\cref{fig:local}).

\subsection{Generative Codebook Sampling}
\label{sec:sampling}
Our goal is to provide means to create and personalize avatars with diverse body shapes, appearances, and local details. To this end, we leverage the above representation to train a \emph{single} multi-subject model which enables the transfer of local features \emph{across} subjects. We note that since the mesh topology of the LBS model is identical, this enables us to learn a shared feature space from multiple posed scans. 

To learn the feature representation over a dataset of $N$ scans, it is sufficient to store the codebooks  $\mathbf{C}_i$ in two  dictionaries $\mathbf{D_s}, \mathbf{D_c}\in \mathbb{R}^{N\times (MF)}$ to represent shape and color information of the $i$-th subject respectively. The entries $\mathbf{C}_i$ can then be learned jointly with the decoder weights via direct 3D supervision using the $i$-th scan (\cref{fig:framework}).
However, we experimentally show that this is insufficient to learn a well-behaved latent space from which we can draw novel samples (see \cref{fig:ablation}). 

Therefore we introduce a codebook sampling strategy that allows us to draw random samples and update the entries of the dictionaries $\mathbf{D_s}, \mathbf{D_c}$ via an additional 2D adversarial loss. 
More specifically, we follow the auto-decoder architecture~\cite{rebain2022lolnerf} and perform PCA on the reshaped dictionary to compute eigenvectors $\mathbf{V}\in \mathbb{R}^{D\times (MF)}$ and fit a normal distribution to the $D$-dimensional PCA coefficients of $N$ samples. A new \emph{random} codebook $\mathbf{C}_r$ can then be generated by sampling $D$-dimensional PCA parameters and multiplying them with the eigenvectors $\mathbf{V}$ (See Supp-B.1 for details). Note that our representation disentangles shapes from appearances with separated geometry and texture branches, which enables independent sampling of geometry and texture features. 



\subsection{Model Training}
\label{sec:training}

\paragraph{3D reconstruction loss.}
To train a codebook $\mathbf{C}_i$ with a single scan, we sample points in a thin shell around the scan. For each coordinate we compute its signed distance $s$ to the input scan, closest texture color $\mathbf{c}$, and surface normal $\mathbf{n}$ on the input scan to attain ground truth values. The codebooks and the decoder weights are then optimized via the following losses: 
\begin{equation}
    \mathcal{L}_{sdf} = \Vert s - s(\mathbf{x_g}) \Vert_{1} + \lambda_n \Vert 1 - \mathbf{n} \cdot \nabla_{\mathbf{x_g}} s(\mathbf{x_g}) \Vert_{1},
\end{equation}
\begin{equation}
    \mathcal{L}_{rgb} = \Vert \mathbf{c} - c(\mathbf{x_g}) \Vert_{1}, 
    \label{eq:rgb}
\end{equation}
\begin{equation}
    \mathcal{L}_{3D} = \lambda_{sdf} \mathcal{L}_{sdf} + \lambda_{rgb} \mathcal{L}_{rgb}. 
\label{eq:3D}
\end{equation}

\paragraph{2D adversarial loss.}

Adversarial learning does not require exact ground-truth annotations but is trained via a collection of real and fake (rendered) images. Thus, real images are obtained by rasterizing the ground-truth scan to which the coarse body mesh $\mathcal{M}$ was fitted. Color and normal images (denoted as ``Real Patch'' in \cref{fig:framework}) are used for learning texture and geometry respectively. Using the same virtual camera parameters and the coarse mesh $\mathcal{M}$, we attain rendered patches (denoted as ``Rendered Patch'' \cref{fig:framework}) via implicit surface rendering of a sampled codebook $\mathbf{C}_r$. Please refer to Supp-B.2 for more details.

Using these 2D patches, we train dictionaries, decoders, and discriminators jointly with a non-saturating logistic loss $\mathcal{L}_{adv}$~\cite{goodfellow2014generative}, R1 regularization $\mathcal{L}_{R1}$~\cite{mescheder2018training}, and path length regularization $\mathcal{L}_{path}$~\cite{karras2020analyzing}. Note that these losses do not require exact ground-truth replication.
Furthermore, we regularize the feature dictionaries to follow a Gaussian distribution~\cite{park2019deepsdf} with $\mathcal{L}_{reg} = \Vert \mathbf{D} \Vert_{F}$.
In summary, we optimize the discriminator: 
\begin{equation}
     \mathcal{L}_{dis} = \mathcal{L}_{adv} + \lambda_{R1} \mathcal{L}_{R1},
\end{equation}
while updating the remaining components (${\mathbf{D_c}, \mathbf{D_s}, \Phi, \Psi}$):
\begin{equation}
     \mathcal{L} = - \mathcal{L}_{adv} + \mathcal{L}_{3D} + \lambda_{path} \mathcal{L}_{path} + \lambda_{reg} \mathcal{L}_{reg},
\end{equation}
where $\lambda_{(\cdot)}$ denotes weights to balance the losses.

Since we sample on the fly during training (see \cref{sec:sampling}), the 2D adversarial loss does affect the shared decoders \emph{and} the whole feature dictionaries (See Supp-Fig.14 for details). 


\subsection{Feature Editing and Avatar Customization}
\label{sec:cutomize}
We now describe how we integrate the above-mentioned human representation and the generative architecture into the avatar creation workflow shown in \cref{fig:teaser}. 
\vspace{-1.em}
\paragraph{Avatar initialization.}
To simplify the avatar creation process, our method allows users to start with a default example $\mathbf{C}_d$, which can be queried from the trained codebook dictionaries directly with index $i$ ($\mathbf{C}_i$) or randomly sampled from the learned D-dimensional PCA parameters distribution ($\mathbf{C}_r$ in \cref{sec:sampling}). 
\vspace{-1.em}
\paragraph{Model fitting.}
Being able to extract elements of interest or copying from existing 3D assets is necessary for avatar creation and editing. To this end, we leverage a similar technique to GAN inversion~\cite{xia2022gan}, where decoder parameters $\mathbf{\Phi}$ and $\mathbf{\Psi}$ are frozen and we only optimize a new feature codebook $\mathbf{C}_{fit}$ to fit a 3D scan.

Given a 3D target scan and its corresponding body model parameters, we calculate the 3D reconstruction loss in \cref{eq:3D} between the prediction conditioned on $\mathbf{C}_{fit}$ and the ground-truth scan, i.e., 
\begin{equation}
    \mathbf{C}_{fit} = \arg \min_{\mathbf{C}} (\lambda_{sdf}\mathcal{L}_{sdf} + \lambda_{rgb} \mathcal{L}_{rgb}). 
    \label{eq:fitting}
\end{equation}
Note that our generative neural fields (MLP decoders) are conditioned on descriptive \emph{local} features. We show that it allows us to accurately fits complex clothing geometry and unseen textural patterns in \cref{sec:fitting_exp}.

\paragraph{Cross-subjects feature editing.}
With multiple codebooks each representing different 3D assets, we can easily transfer local geometry and texture from one avatar to another, for instance, changing the top wear from the fitted scan $\mathbf{C}_{fit}$ to the initial $\mathbf{C}_d$.
Recalling that all codebooks are indexed by an identical mesh topology, users can easily retrieve the index numbers of  $\mathcal{V}_{body} \subset \mathcal{V}$ via standard mesh visualization tools such as Blender~\cite{blender}. Finally, swapping of the corresponding rows in $\mathbf{C}_{fit}$ and $\mathbf{C}_d$ given the vertex indices also swaps the local appearance. 
\begin{figure}[t]
\centering
\includegraphics[width=\linewidth]{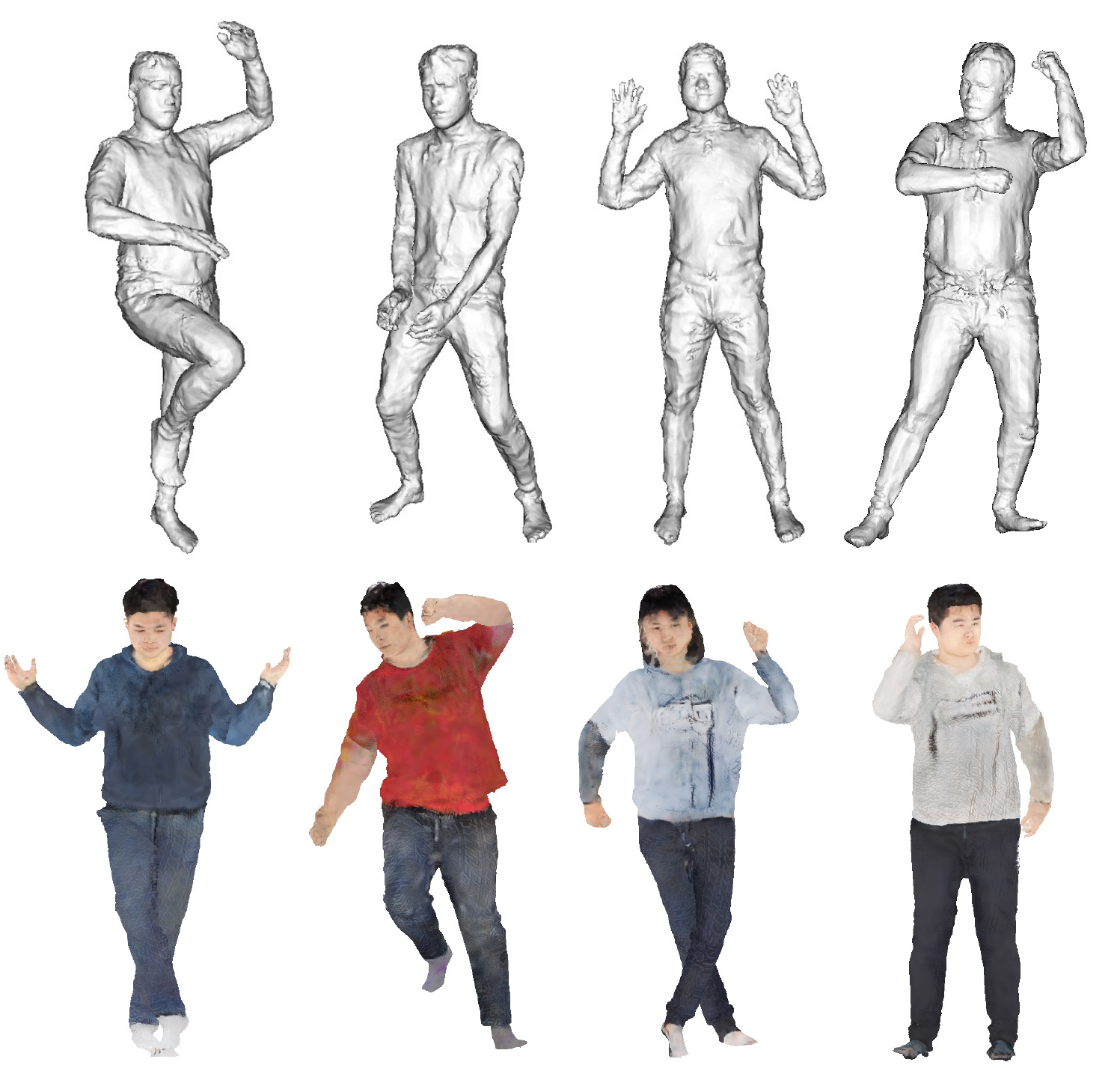}
\caption{\textbf{Randomly sampled texture and body geometry from the model trained on THuman2.0}. \emph{Top}: Given any poses, our randomly sampled geometries contain realistic details such as wrinkles in garments and facial expressions.
\emph{Bottom}: Given arbitrary poses and body geometries, our model produces reasonable colors on skin, hair, clothes, and pants in each sample (We turn off the shader for visualizing pure texture colors).}
\label{fig:random}
\vspace{-1.5em}
\end{figure}
\vspace{-1.em}
\paragraph{Personalized texture drawing.} 
One can further customize clothing by drawing directly onto 2D images. Due to the learned disentangled feature spaces, we are able to update texture features in $\mathbf{C}_{fit}$ while keeping the geometry features unchanged. When fitting 2D images, users draw on the rasterized images from arbitrary target scans. We then finetune only texture features in $\mathbf{C}_{fit}$ via the RGB loss in \cref{eq:rgb} given the corresponding 3D coordinates and colors.
\vspace{-2em}
\paragraph{Avatar reposing.}
Our representation combines a base mesh $\mathcal{M}$ and underlying feature codebooks $\mathbf{C}_{fit}$ that learn \emph{local} geometry and texture on the 3D scans. Hence, one can repose the mesh $\mathcal{M}$ using ($\theta, \beta$) parameters, which also reposes the avatar correspondingly. Since the geometry codebook does not contain global pose ($\theta, \beta$), local information can be consistently applied to $\mathcal{M}$ under unseen poses.

\section{Experiments}
\begin{figure}[t]
\centering
\includegraphics[width=\linewidth]{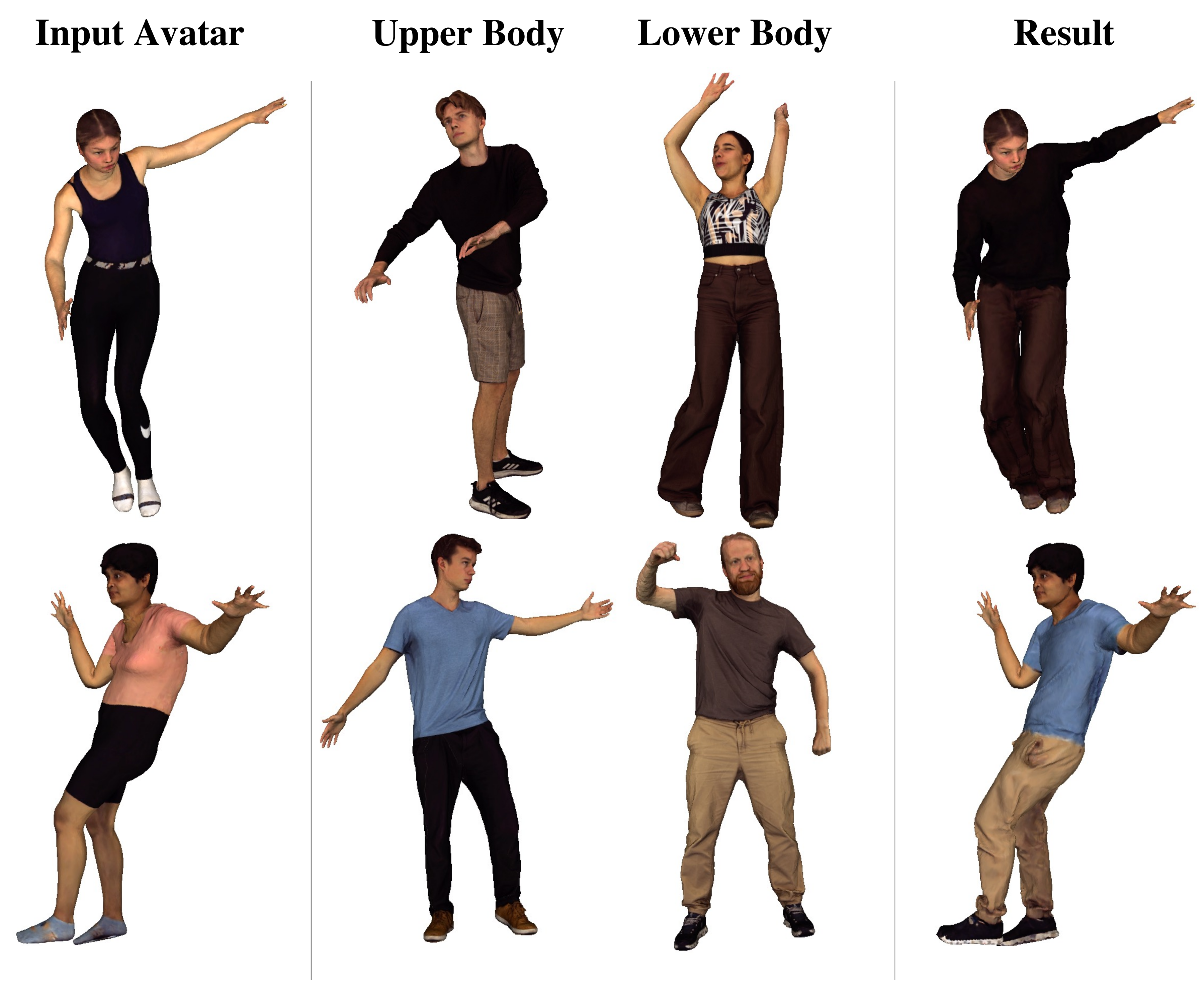}
\caption{\textbf{Cross-subject feature editing results}. We  partially transfer local clothing details from the unseen scans (upper and lower body) to the input avatars. The results of the edited avatars are shown in the right column. }
\label{fig:editing}
\vspace{-1.5em}
\end{figure}

Our goal is locally editable 3D avatar creation. Since we are the first to discuss this problem, we visualize our editing results in \cref{sec:avatar_exp}. Next, we highlight the capability of model fitting by comparing our method with SOTA human generative models in~\cref{sec:fitting_exp}. Finally, controlled experiments are presented in \cref{sec:ablation_exp} and \cref{sec:data_exp} to verify the effectiveness of our design.


\subsection{Experiment Settings}
\label{sec:dataset_exp}

\paragraph{Dataset.}
Most generative human works~\cite{chen2022gdna,palafox2021npms,palafox2021spams} exploit commercial data~\cite{3dpeople,renderpeople} for training, which is not easily accessible and limits reproducibility. Furthermore, the quality of publicly available 3D human datasets~\cite{tiwari20sizer,tao2021function4d} is not satisfactory. Issues such as non-watertight topologies and noise are very common (See Supp-A.2 for examples and comparison). To bridge this gap, we collect a new dataset named \textbf{CustomHumans} for training and evaluation. Here we summarize the datasets used in our experiments.
\begin{itemize}
    \item \textbf{CustomHumans} (Ours) contains more than 600 high-quality scans of 80 participants in 120 garments in varied poses from a volumetric capture stage~\cite{collet2015high}, which is equipped with 106 synchronized cameras (53 RGB and 53 IR cameras). We use our dataset to train models of all quantitative experiments. (\cref{sec:fitting_exp} $\sim$~\cref{sec:data_exp})
    \item \textbf{THuman2.0}~\cite{tao2021function4d} is a dataset containing about 500 scans of humans wearing 150 garments in various poses. Since this dataset has more textural diversity, we train our method on it for qualitative random sampling experiments (\cref{sec:avatar_exp} and~\cref{sec:ablation_exp}).
    \item \textbf{SIZER}~\cite{tiwari20sizer} is a widely used 3D scan dataset containing A-pose human meshes of 97 subjects in 22 garments. These meshes are used as \emph{unseen} test scans in our fitting experiment (\cref{sec:fitting_exp}).
\end{itemize}
\paragraph{Evaluation protocol.} Following the evaluation protocol in OccNet~\cite{mescheder2019occupancy}, we quantitatively evaluate the model fitting accuracy using three metrics: \textbf{Chamfer distance (CD)}, \textbf{normal consistency (NC)}, and \textbf{f-Score}.
\begin{figure}[t]
\centering
\includegraphics[width=\linewidth]{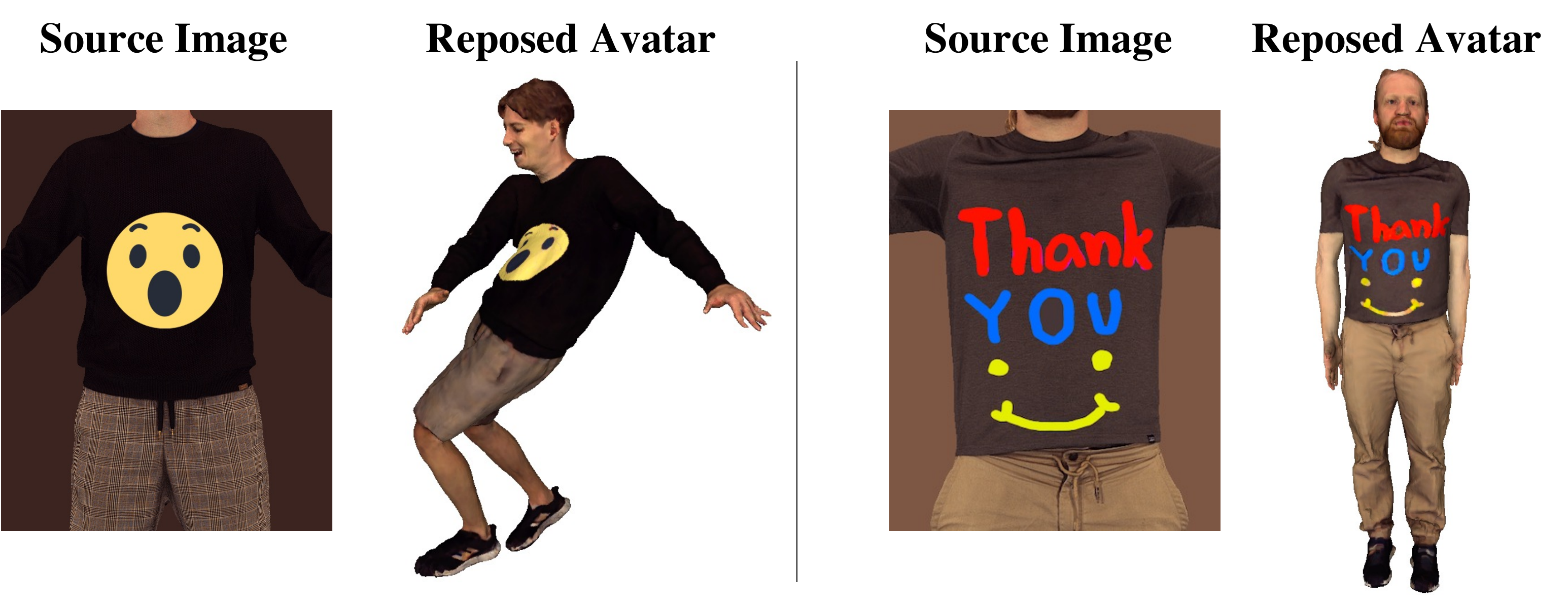}
\vspace{-2em}
\caption{\textbf{Personalized texture editing}. We draw personalized logos on 2D images and fit avatars' texture features to the images. These local textures remain consistent under pose changes.}
\label{fig:drawing}
\vspace{-1.5em}
\end{figure}

\subsection{Customized Avatars}
\label{sec:avatar_exp}
We visualize the results of our proposed avatar customization workflow described in \cref{sec:cutomize}.
\vspace{-1.em}
\paragraph{Avatar initialization.} In \cref{fig:random}, we show random textures and geometries sampled from the model trained on THuman2.0. Our method is able to generate reasonable colors and wrinkles in arbitrary poses. Note that the sampled geometries are shown as the \emph{real} meshes but not as rendered normals as in~\cite{chen2022gdna} (See Supp-C.2 for comparisons). 
\vspace{-1.em}
\paragraph{Cross-subjects feature editing.}
After fitting feature codebooks to 3D scans, we can change the clothes on our avatars by swapping the local features stored on the body vertices.
We select the features within the upper body and lower body areas. We then copy these local features to the initial avatars’ feature codebooks. As shown in \cref{fig:editing}, our method is able to handle multiple garments on different human subjects and preserves consistent details under different body poses or shapes. 
\vspace{-1.em}
\paragraph{Personalized texture drawing.}
Our method allows users to draw complex letters and logos on images for personalized texture editing. We perform the model fitting and feature editing process but only optimize the texture features in the codebooks using user-edited images and the RGB loss (\cref{eq:rgb}). \cref{fig:drawing} shows that new texture can be seamlessly applied to the 3D avatars. It is worth noting that resulting avatars enable detailed pose control via the SMPL-X parameters without affecting the fitted texture and geometry. 
\begin{figure}[t]
\centering
\includegraphics[width=\linewidth]{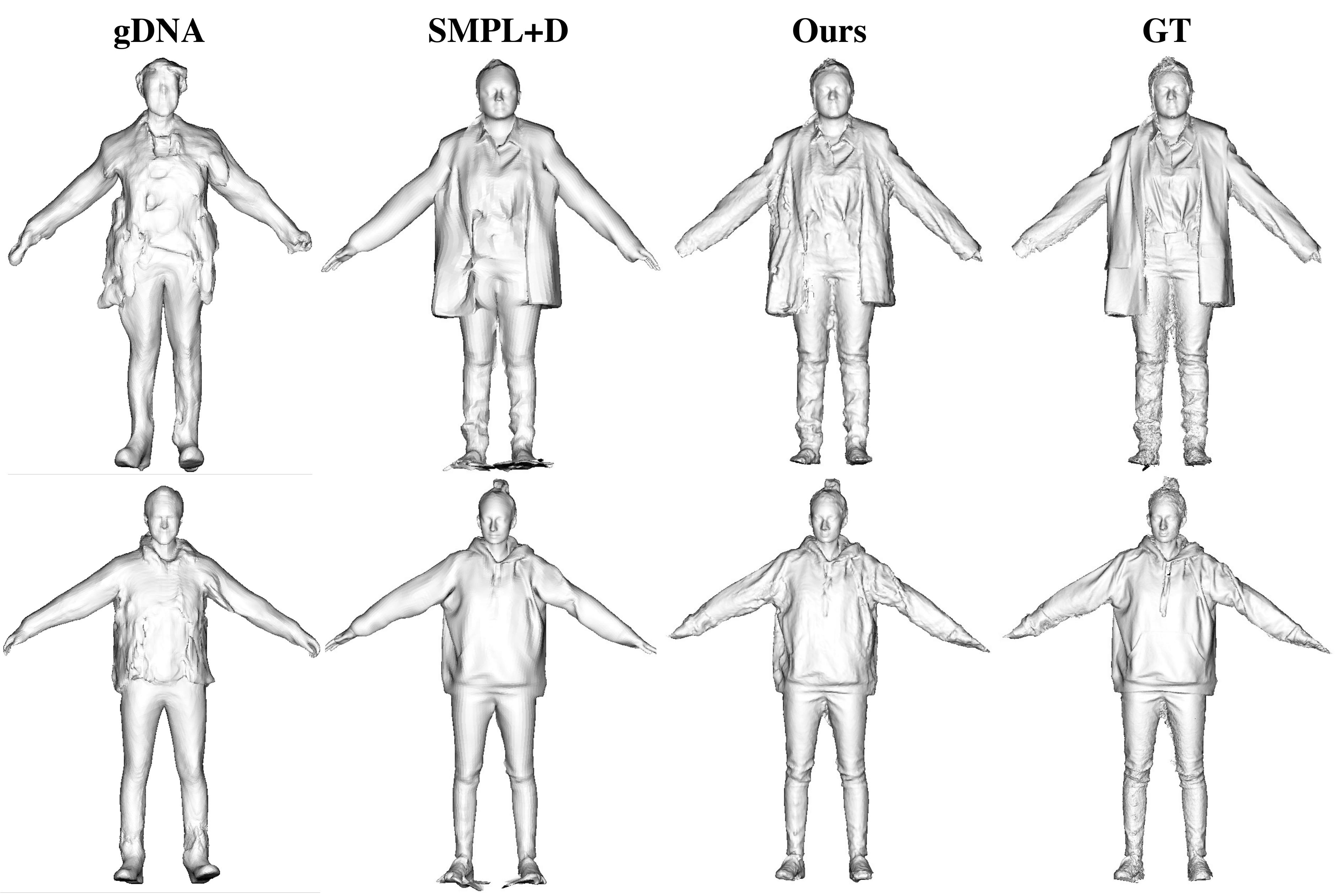}
\caption{\textbf{Qualitative comparison of model fitting on SIZER}. We visualize the fitting results from gDNA~\cite{chen2022gdna}, SMPL+D~\cite{alldieck2019learning}, and our method. Our results are perceptually close to the ground truth even on the challenging test cases of jackets and loose t-shirts.}
\label{fig:fitting}
\end{figure}

\begin{figure}[t]
\centering
\includegraphics[width=0.8\linewidth]{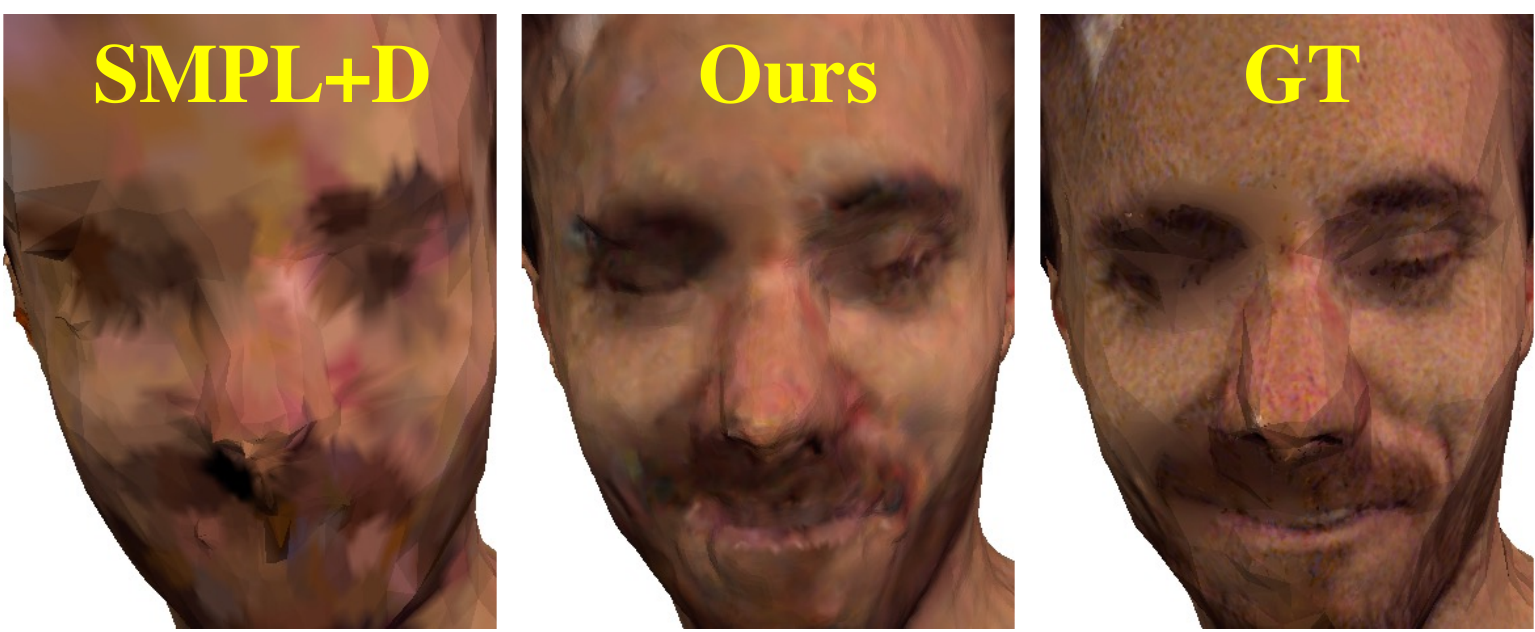}
\caption{\textbf{Qualitative comparison of texture fitting}. We compare our method with SMPL+D~\cite{alldieck2019learning} by fitting to unseen textured meshes. The performance of SMPL+D is limited by its geometry and texture resolution.}
\label{fig:manface}
\vspace{-1.5em}
\end{figure}

\vspace{-1.em}
\subsection{Model Fitting Comparison}
\label{sec:fitting_exp}
Since model fitting is an important step in our avatar creation workflow, we compare the capability of feature inversion using unseen 3D scans. The goal of this task is to invert a 3D scan into latent codes while keeping the remaining model parameters fixed. We compare our method with the 3D human generative model gDNA~\cite{chen2022gdna}, which has achieved state-of-the-art performance in fitting the geometry of 3D human bodies.
We also directly compare with SMPL~\cite{SMPL:2015} and SMPL+D~\cite{alldieck2019learning}. Note that SMPL+D is a stronger vertex-based extension that uses a subdivided version of SMPL to directly register surfaces to scans while our method and gDNA optimize latent codes.

From \cref{fig:fitting} we can see that SMPL+D handles loose clothing, such as a business suite, better than gDNA. However, the surfaces of SMPL+D results are over-smoothed and do not contain high-frequency details while ours can preserve them. Quantitatively, our method consistently outperforms these methods on all metrics as shown in \cref{tab:fitting}.

\cref{fig:manface} depicts the result of texture fitting against the SMPL+D baseline. While both methods inherit a fixed mesh topology the quality of SMPL+D is limited by its model resolution. Our method addresses this issue via local neural fields that enable cross-subject feature editing of texture and geometry with enhanced representational power.

\subsection{Ablation Study}
\label{sec:ablation_exp}
\begin{figure}[t]
\centering
\includegraphics[width=\linewidth]{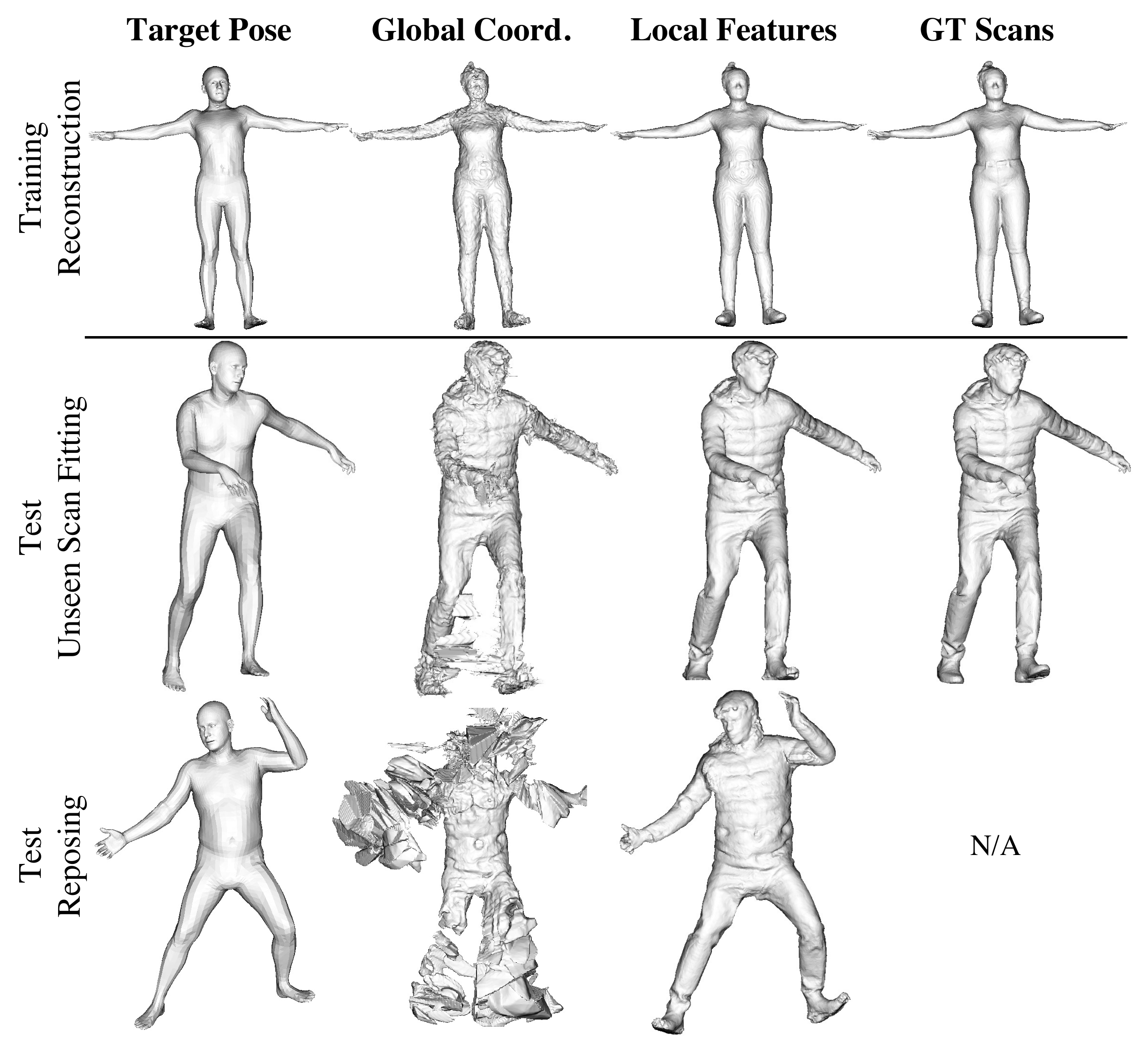}
\caption{\textbf{Comparison of using global/local features for conditioning decoders}. The use of global coordinates causes overfitting and memorization to the shared decoders, which makes it struggle to handle unseen scans or novel body poses.}
\label{fig:local}
\end{figure}

%
%
\begin{table}[t]
    \centering
    \small
\begin{tabular}{lccc}
\toprule
 Method               & \begin{tabular}[c]{@{}c@{}}Pred-to-Scan /\\ Scan-to-Pred (mm)$\downarrow$ \end{tabular} & NC$\uparrow$ & f-Score$\uparrow$     \\
\midrule
 SMPL~\cite{SMPL:2015}         & 13.60 / 18.03  & 0.849 & 0.458        \\
 gDNA~\cite{chen2022gdna}         & 8.374 / 8.006  &	0.842 & 0.718        \\
 SMPL+D~\cite{alldieck2019learning}       & 5.192 / 2.854 & 0.911 & 0.962        \\
 Ours                 & \textbf{1.364} /\textbf{ 1.423}  & \textbf{0.949} & \textbf{0.997}        \\
\bottomrule
\end{tabular}
\caption{\textbf{Model fitting comparison on SIZER}. We report Chamfer distance, normal consistency (NC), and f-score between ground truth and the meshes fitted by different methods.}
\vspace{-1.5em}

\label{tab:fitting}
\end{table}


\begin{table*}[t]
    \centering
\begin{tabular}{c|ccccc}
\toprule
 Training Data Percentage     &  10\%   & 25\%    & 50 \%    & 75\%      & 100\%          \\
\midrule
  \begin{tabular}[c]{@{}c@{}}Chamfer Distance (mm) \\ S-to-P / P-to-S $\downarrow$ \end{tabular}   &  1.933 / 1.798 &  1.754 / 1.590 & 1.543 / 1.456 & 1.463 / 1.385 &  1.423 / 1.364 \\
Normal Consistency$ \uparrow$          & 0.918  &  0.931   &  0.935   &  0.947    & 0.949          \\
f-Score (\%)  $\uparrow$           & 99.25  &  99.38   &  99.65   &  99.74    & 99.75          \\
\bottomrule
\end{tabular}
\caption{\textbf{Generalization analysis on CustomHumans}.  We analyze the model fitting performances with regard to different amounts of training data (100\% = 100 training scans). We observe consistent performance gain on all evaluation metrics when using more training subjects to train the shared decoders. }
\label{tab:data}
\vspace{-2em}
\end{table*}

%

\begin{figure}[t]
\centering
\includegraphics[width=\linewidth]{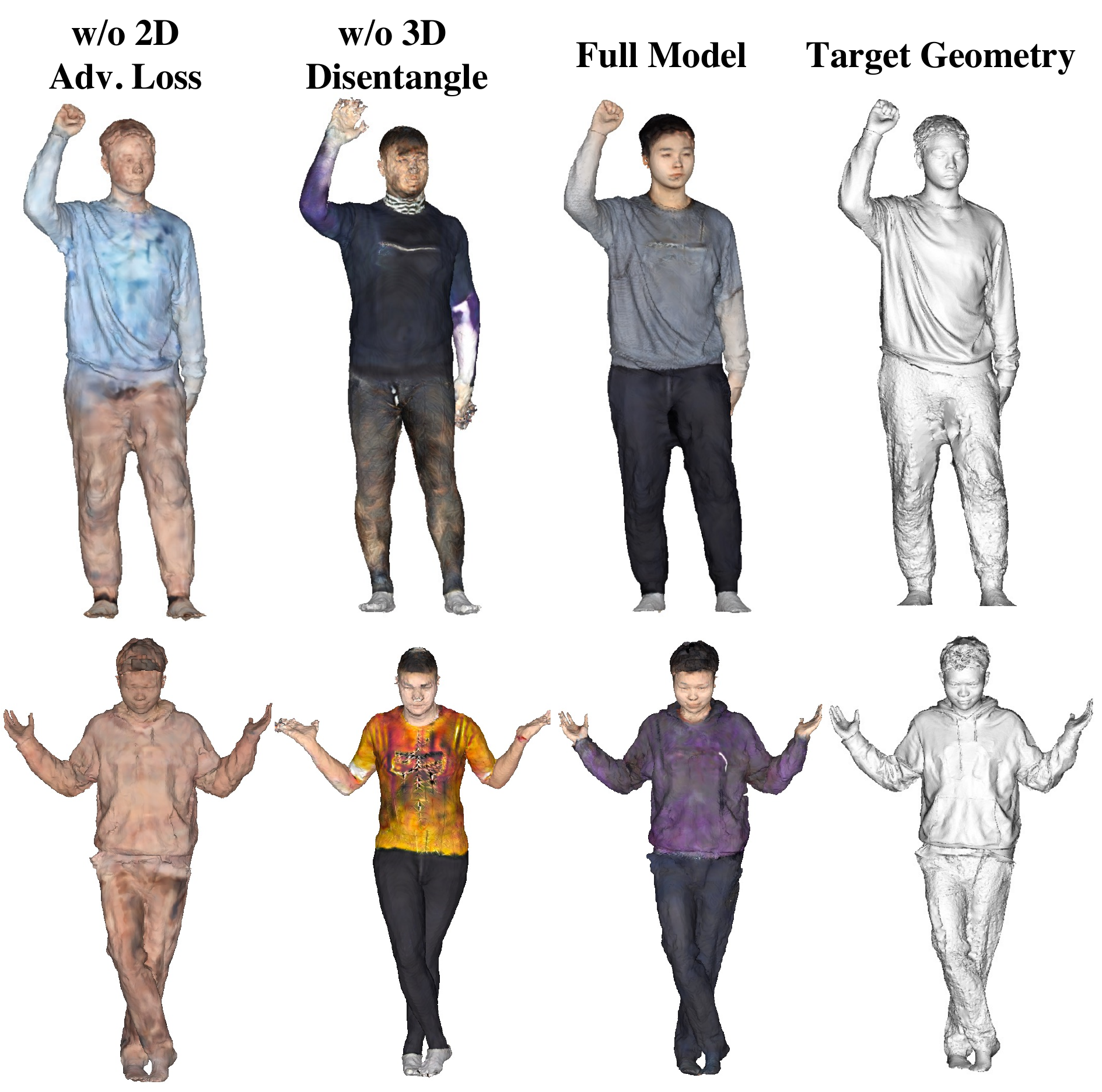}
\caption{\textbf{Ablative comparison of our framework designs}. We visualize the results of transferring random texture to given body geometry. Our full model produces reasonable body texture and is able to maintain fixed geometry for texture editing.}
\label{fig:ablation}
\vspace{-1.5em}
\end{figure}

\paragraph{Effectiveness of local features and shared decoders.}
To verify the design choice of using local features, we replace the local features $\mathbf{x}_l$ by the global coordinates $\mathbf{x}_g$ for conditioning the decoders. \cref{fig:local} shows that even though the shared decoders are able to achieve similar reconstruction results when training with global information, they do not maintain consistent performance for model fitting and avatar reposing. This is because the shared decoders tend to memorize global coordinates information in a ``per-subject'' manner, rather than learning shareable information that can be used across vertices and subjects. On the other hand, our representation ensures only local features defined on the triangle coordinates are exposed to the shared decoders. In such cases, the decoders can better handle unseen body poses or out-of-distribution samples for model fitting and avatar editing.
\vspace{-1.em}
\paragraph{Importance of 2D adversarial loss and 3D disentanglement.}
As discussed in \cref{sec:training}, we introduce feature disentanglement and generative adversarial learning in our training framework. As shown in \cref{fig:ablation}, sampling within the feature spaces learned without adversarial loss does not yield reasonable body textures. Similarly, training only a single decoder for both geometry and texture does not allow us to maintain desired body geometries when sampling random textures. Our full model can produce disentangled textures, given arbitrary body geometries and poses.


\subsection{Generalization Ability Analysis}
\label{sec:data_exp}

\begin{figure}[t]
\centering
\includegraphics[width=\linewidth]{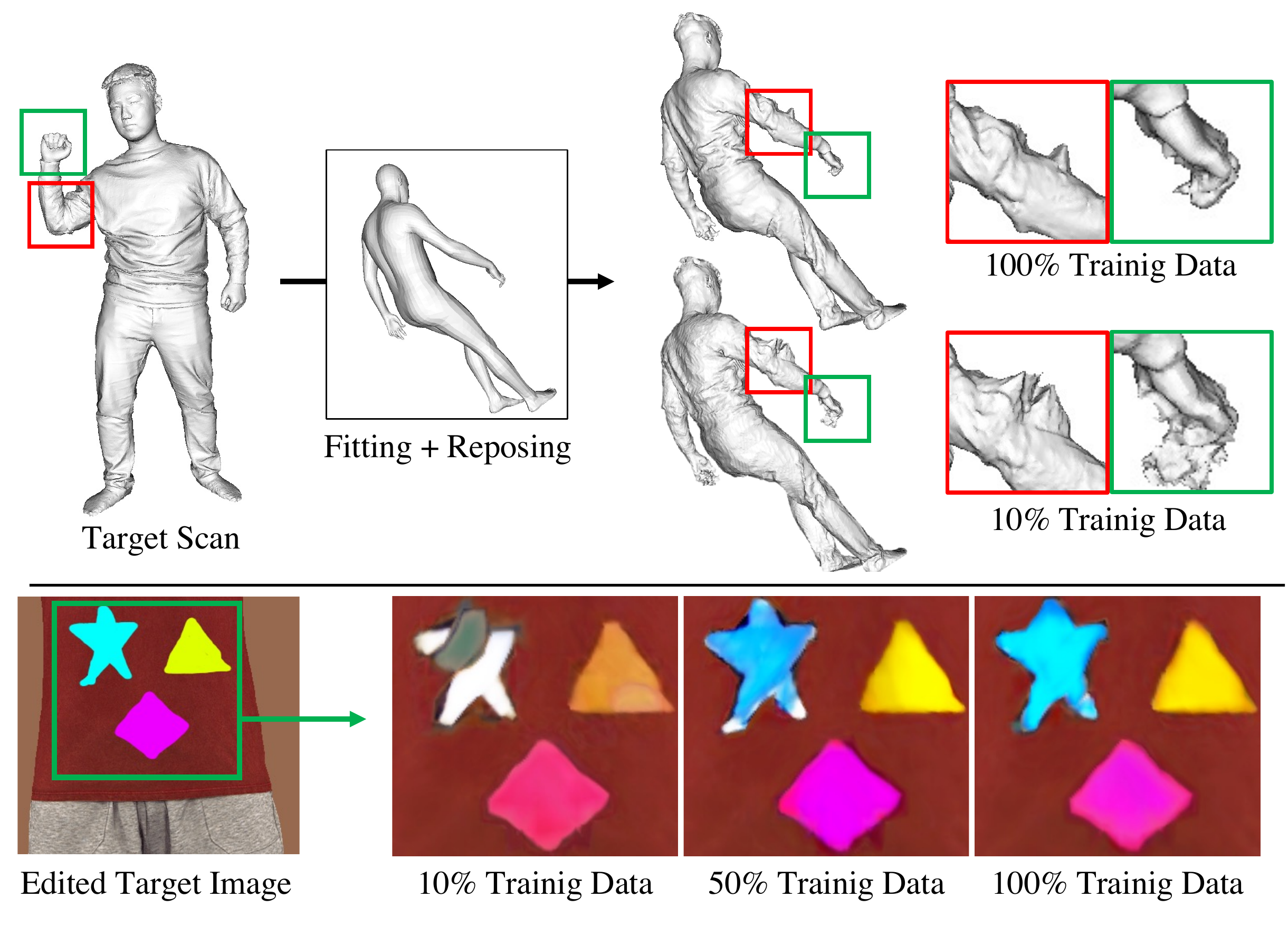}
\caption{\textbf{Qualitative comparison of generalization}. \emph{Top}: We visualize the results of avatar reposing using different percentages of training data (i.e., 10 meshes vs 100 meshes). Artifacts caused by self-contact (e.g., fist and elbow) can be reduced when more training subjects are introduced. \emph{Bottom}: We visualize the results using different percentages of training data. Using more data results in more robust shared decoders (with a wider color range), which is necessary for the avatar creation task.}
\label{fig:data_repose}
\vspace{-1em}
\end{figure}

We are interested in how the amount of training data affects the capacities of the MLP decoders. To analyze this, we design three evaluation protocols: 3D model fitting, avatar reposing, and 2D texture fitting. \cref{tab:data} summarizes the model fitting performance using different percentages of training data. We observe a $25\%$ accuracy improvement when using the full training set. In \cref{fig:data_repose} (\emph{Top}) we show that the reposing artifacts caused by
self-contact (e.g., fist and elbow) can be reduced when training the MLP decoders with more poses and subjects. In addition, \cref{fig:data_repose} (\emph{Bottom}) depicts a qualitative comparison of 2D texture editing under different training data percentages. We evaluate texture editing quality by fitting a 2D image with unseen geometric shapes and colors. It can be seen that the model trained on more samples is able to handle a wider range of color distribution. These results confirm the necessity for learning multi-subject shared decoders in our task.

\section{Conclusion}

We propose an end-to-end trainable framework for learning 3D human avatars with high fidelity and full editability.
By combining neural fields with explicit skinned meshes, our representation addresses the controllability issue of many previous implicit representations.
Moreover, we uniquely integrate the proposed human representation into a generative auto-decoding pipeline that enables local editing across \emph{multiple} animation-ready avatars. 
Through our evaluation on the newly contributed CustomHumans dataset, we demonstrate that our approach achieves higher model fitting accuracy and generates diverse detailed avatars.
We believe that this work opens up exciting possibilities for accelerating content creation in the Metaverse.
\vspace{-1.em}
\paragraph{Acknowledgements} We express our gratitude to Stefan Walter and Dean Bakker for infrastructure support, Juan Zarate for managing the capture stage, and Deniz Yildiz and Laura Wülfroth for data capture assistance. We thank Andrew Searle for supporting the capturing system and all the dataset participants.

{\small
\bibliographystyle{ieee_fullname}
\bibliography{egbib}
}

\end{document}


\title{Supplementary Material for \\
``Learning Locally Editable Virtual Humans”}

\author{Hsuan-I Ho
\and
Lixin Xue
\and
Jie Song
\and
Otmar Hilliges
\and
Department of Computer Science, ETH Zürich}

\maketitle


{
  \hypersetup{linkcolor=black}
  \tableofcontents
}

\noindent\rule{\linewidth}{0.4pt}


\section{CustomHumans Dataset}
\subsection{Dataset description}

\begin{figure}[t]
\centering
\includegraphics[width=\linewidth]{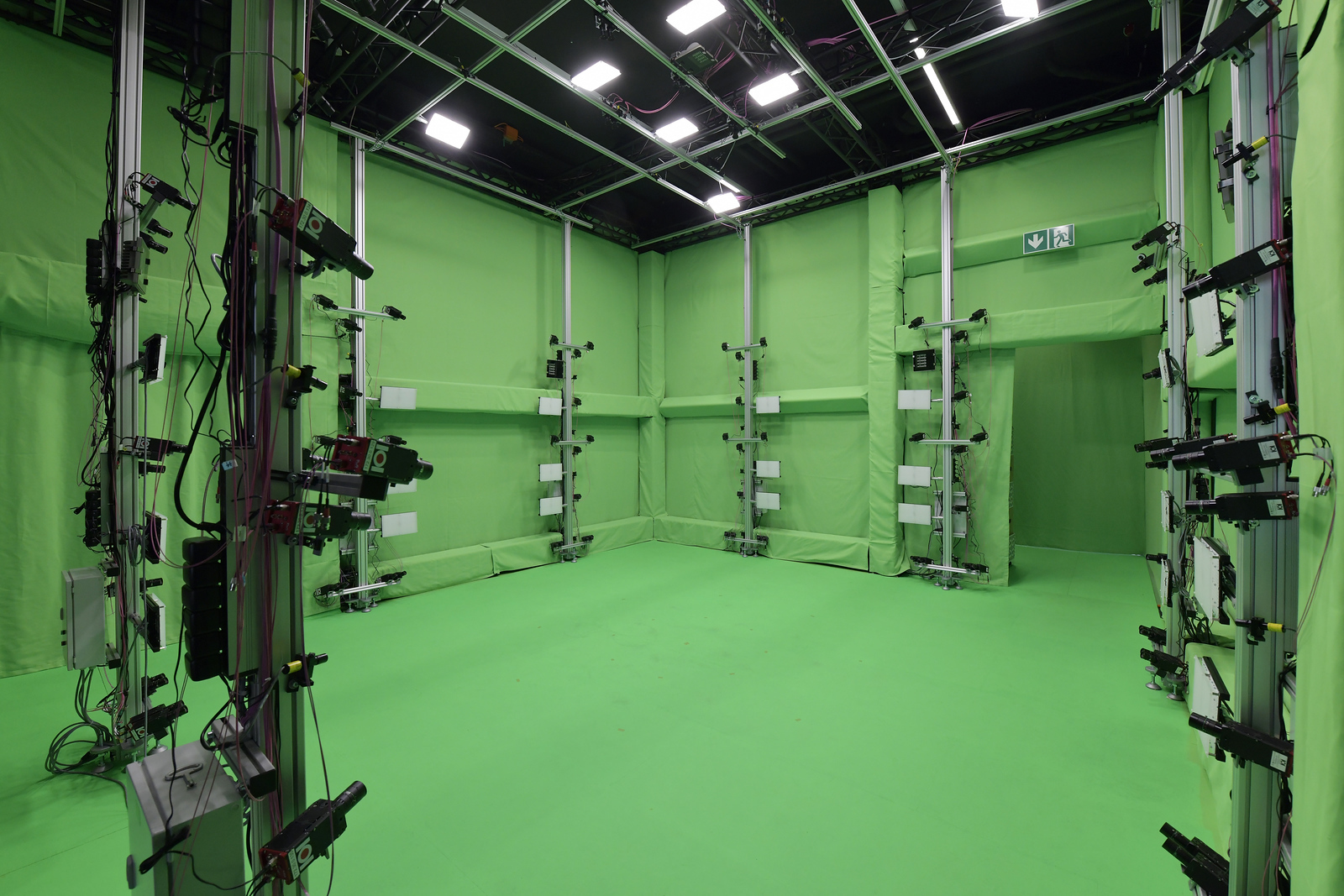}
\caption{\textbf{Volumetric capture stage for data collection}. Our capture stage is equipped with 106 synchronized cameras (53 RGB and 53 IR cameras) for capturing dynamic 4D sequences.}
\label{fig:capture}
\end{figure}

In this section, we provide more information about our contributed dataset, CustomHumans. Our dataset is recorded by a multi-view photogrammetry system~\cite{collet2015high} as shown in \cref{fig:capture}, equipped with 53 RGB (12 Megapixels) and 53 (4 Megapixels) IR cameras. The resulting high-quality scan is composed of a 40K-face mesh alongside a 4K-resolution texture map. In addition to the high-quality scans, we provide accurately registered SMPL-X parameters using a customized mesh registration pipeline.

\begin{figure*}[t]
\centering
\includegraphics[width=0.8\linewidth]{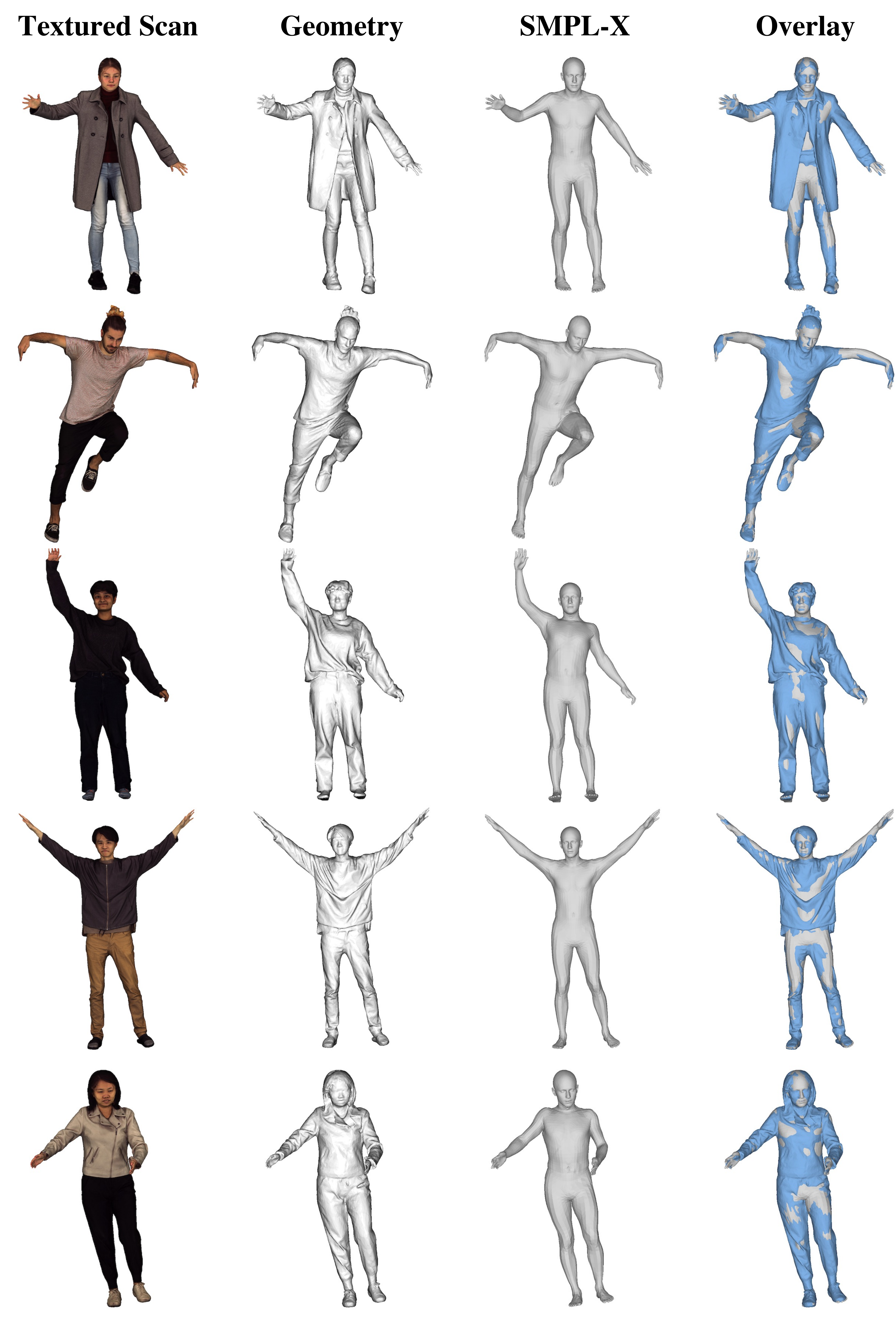}
\caption{\textbf{Visualization of data samples in CustomHumans}. In our dataset, we provide high-quality human meshes, high-resolution texture maps, and registered SMPL-X body models. To verify the accuracy of registration, we visualize the overlays of the human scans and the SMPL-X body meshes.  }
\label{fig:dataset}
\end{figure*}

To collect clothed human scans in various poses, we invited 80 participants to our capture studio. We designed several movement instructions for the participants, such as ``T-pose'', ``Hands Up'', ``Squat'', ``Turing head'', and ``Hand gestures'', to film 5-6 poses in a 10-second long sequence (300 frames). We selected 4-5 best-quality meshes in each sequence as our data samples. In total, our dataset contains more than 600 high-quality scans with 120 different garments. Exemplars of human scans can be found in~\cref{fig:dataset}, where we visualize the textured scans, mesh geometries, and the registered SMPL-X body models.

\subsection{Comparison with existing datasets}

\definecolor{Gray}{gray}{0.85}

\begin{table*}[ht]
    \centering
\begin{tabular}{lccccccc}
\toprule
  Dataset         &  \begin{tabular}[c]{@{}c@{}} Subject \\ Diversity \end{tabular} &
                     \begin{tabular}[c]{@{}c@{}} Garment \\ Diversity \end{tabular} & 
                     \begin{tabular}[c]{@{}c@{}} Pose \\ Variation \end{tabular}    &
                     Noise-free  &  Watertight & Registered &
                     \begin{tabular}[c]{@{}c@{}} Publicly \\ Available \end{tabular}       \\
\midrule

 RenderPeople~\cite{renderpeople}       &  \checkmark    &  \checkmark     & \checkmark   & \checkmark     &  \checkmark  &    &    \\
 \rowcolor{Gray}
 CAPE~\cite{CAPE:CVPR:20}           &       &        & \checkmark   & \checkmark     &  \checkmark  & \checkmark  & \checkmark   \\
 SIZER~\cite{tiwari20sizer}                 &  \checkmark    &        &      &      &     & \checkmark  & \checkmark  \\
 \rowcolor{Gray}
 THuman2.0~\cite{tao2021function4d}             &       &  \checkmark     & \checkmark    &      &  \checkmark  & \checkmark  & \checkmark  \\
 \midrule
 Ours                  &  \checkmark    &  \checkmark     & \checkmark    &  \checkmark   &  \checkmark  & \checkmark  & \checkmark  \\
\bottomrule
\end{tabular}
\caption{\textbf{Comparison with existing 3D human scans datasets}. Commercial datasets such as RenderPeople are not easy to obtain whereas either the quality (e.g. SIZER and THuman2.0) or diversity (e.g., CAPE) is not sufficient in the other datasets.}
\label{tab:dataset}
\end{table*}


We summarize the outstanding features of existing 3D clothed human datasets in  \cref{tab:dataset}. Specifically, we are mainly interested in four aspects. \textbf{Subject Diversity}: Does it contain people of diverse genders and races?  \textbf{Garment Diversity}: Does it include various clothing and combinations? \textbf{Pose Variation}: Does it consists of subjects in various poses? \textbf{Quality}: Do these scans contain noise near the surfaces? and are they watertight? 
\begin{figure*}[htbp]
\centering
\includegraphics[width=\linewidth]{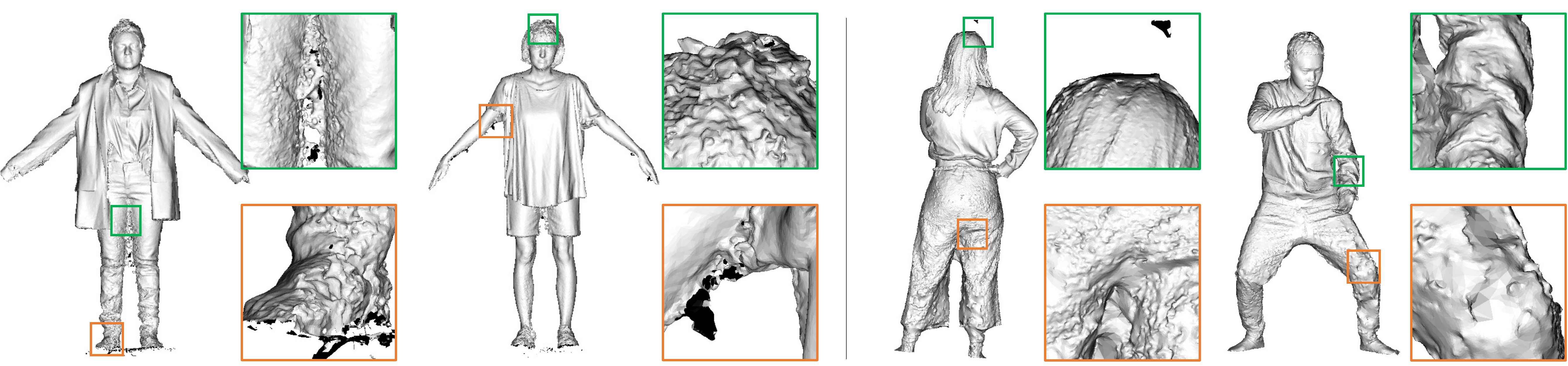}
\caption{\textbf{Examples of noisy data in SIZER~\cite{tiwari20sizer} (left) and THuman2.0~\cite{tao2021function4d} (right)}. The meshes in SIZER and THuman2.0 are generally not watertight and bumpy, which causes issues when learning detailed human body geometry. }
\label{fig:noisy}
\end{figure*}

Commercial datasets (e.g., RenderPeople~\cite{renderpeople}) have shown superior quality and have been used in many works of generative modeling. However, they are not easily accessible which limits reproducibility for research purposes. CAPE~\cite{CAPE:CVPR:20} contains posed sequences of 15 subjects and 8 types of outfits. Since only SMPL+D body meshes are used in this dataset, they are clean and watertight. The total number of subjects and garments is limited as the main focus of this dataset is to model pose-dependent deformations of a single subject.
SIZER~\cite{tiwari20sizer} provides 2000 scans of 100 different subjects in a total of 22 types of garments. Each scan is captured in ``A-Pose'' and registered by SMPL and SMPL+D body models. Nevertheless, as shown in~\cref{fig:noisy}~\emph{left}, these scans contain non-watertight mesh manifolds and large noise near surfaces. 
THuman2.0~\cite{tao2021function4d} consists of 525 scans of approximately 150 subjects and garments captured by a dense DLSR rig, which limits the mesh quality due to noise and bumpy surfaces (\cref{fig:noisy}~\emph{right}). 
To foster future research on creating detailed human avatars, we addressed the above-mentioned issues and collected a new high-quality dataset containing 600 posed scans with higher subject diversity and in various clothing.

\section{Implementation Details}

\subsection{Codebook sampling}
\begin{figure*}[t]
\centering
\includegraphics[width=0.8\linewidth]{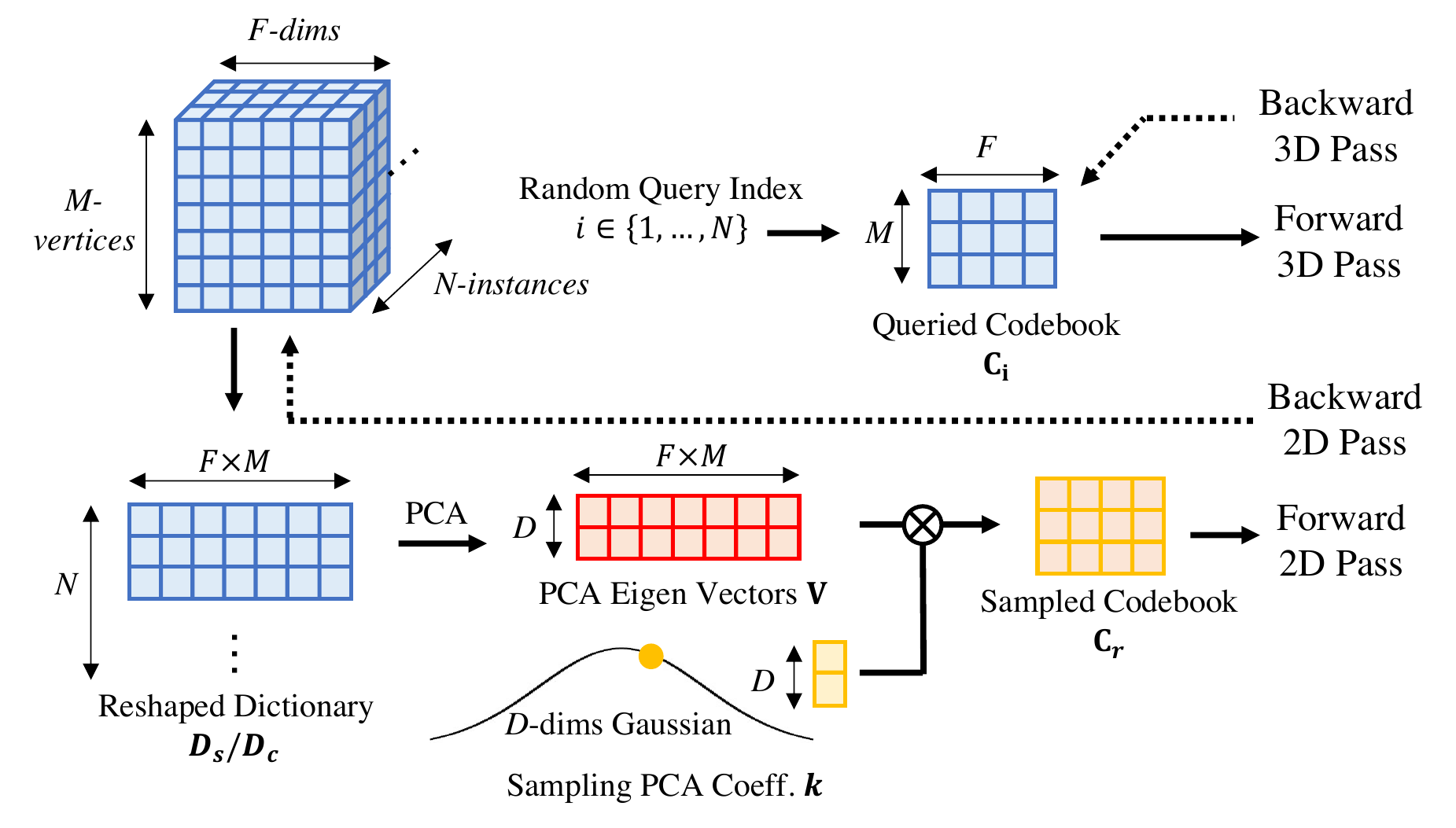}
\caption{\textbf{Two codebook sampling strategies for training.} \emph{Top}: Given a random subject index $i$, we query the corresponding codebook stored in the dictionary. This \textbf{queried codebook} $\mathbf{C}_i$ is trained via 3D supervision, and the gradients only back-propagate to the decoder and $\mathbf{C}_i$. \emph{Bottom}: We apply PCA to the high-dimensional dictionary, and draw random coefficients $\mathbf{k}$ from the $D$-dimensional PCA space. As the \textbf{sampled codebook} $\mathbf{C}_r$ is a linear combination of the PCA eigenvectors $\mathbf{V}$, the 2D adversarial loss can be used for updating the whole dictionary. Note that both strategies are applied during training to jointly optimize both the dictionaries and the decoders. }
\label{fig:sampling}
\vspace{-1em}
\end{figure*}
\cref{fig:sampling} depicts the codebook sampling strategies used for training our model. As mentioned in Sec. 3.2, we store the codebooks $\mathbf{C}_i$ in two dictionaries $\mathbf{D_s}, \mathbf{D_c}\in \mathbb{R}^{N\times (MF)}$ to represent shape and color information of the $i$-th subject. This allows us to learn a unified shared feature space and transfer local feature codebooks across subjects. The entry $\mathbf{C}_i$ is queried to be jointly trained with the decoder weights via direct 3D supervision.

To further learn a well-behaved latent space from which we can draw novel samples, we devise an on-the-fly PCA codebook sampling strategy inspired by~\cite{rebain2022lolnerf}. We first compute $D$-dimensional eigenvectors $\mathbf{V}\in \mathbb{R}^{D\times (MF)}$ by applying PCA to the reshaped dictionary. We then derive the $D$-dimensional PCA coefficients of all $N$ samples using the eigenvectors $\mathbf{V}$. The mean and the covariance matrix of these PCA coefficients can be computed and used for fitting a $D$-dimensional normal distribution. We then draw random PCA coefficients $\mathbf{k}$ from the normal distribution and compute a new codebook $\mathbf{C}_r$ by multiplying $\mathbf{k}$ and $\mathbf{V}$.

As shown in \cref{fig:sampling}, the 3D loss is used for updating only the selected codebook $\mathbf{C}_i$. On the other hand, the 2D adversarial loss can be backpropagated to the entire dictionary since the on-the-fly PCA operation is differentiable. This PCA sampling strategy allows us to learn a more meaningful latent space for all training samples instead of overfitting each sample independently. 

\subsection{Implicit rendering}

As described in Sec. 3.3, we render local color and normal patches for adversarial learning by rasterizing the ground-truth scans. To do so, we place the hip joint of each human scan at the origin and place 4 virtual cameras on $\{0, 90, 180, 270\} \degree$ of a 2-meter circle. We rasterize images of the full body in $1024 \times 1024$ and then crop each image to 25 $128 \times 128$ patches based on the body joint positions defined on the SMPL-X model.

We then use the same virtual camera parameters to shoot rays (i.e., pixels on the image patches) from the camera center onto the implicit surface.
Sample points on a camera ray can be formulated as $ \mathbf{x} = \mathbf{r}_o + t \times \mathbf{r}_d $, where $\mathbf{r}_o$ is the ray origin, $\mathbf{r}_d$ is the ray direction and $t$ is a scalar for sampling. 
We determine the intersection by finding the first SDF sign-changing sample along the ray following~\cite{chen2022gdna}.
These intersection coordinates will be the query points for the feature querying and decoding process to predict corresponding colors and SDF. We compute the finite differences of SDF as an approximation of surface normals. The resulting normal and color maps are served as ``fake image patches'' for the discriminators.

\subsection{Network architecture}

We choose SMPL-X~\cite{pavlakos2019expressive} as our LBS body mesh, which consists of $M=10475$ vertices. We use a feature dimension of $F=32$ for both texture and geometry features, resulting in a codebook of $10475 \times 64$ for each subject. A positional encoding of 5 frequency bands is applied to the local positional features $\mathbf{x}_l$. Our shared decoders consist of 4 layers of 128-dimensional linear layer followed by a ReLU activation. Our discriminator follows similar network architecture with StyleGAN2~\cite{karras2020analyzing}. We apply two different discriminators for normal maps and color images.

\subsection{Training details}
We use a dictionary size of $N=150$ for THuman2.0~\cite{tao2021function4d} and $N=100$ for the CustomHumans dataset. For better facial textures and finger control, we register both datasets with SMPL-X parameters $\theta, \beta$ including facial contours and finger joints.

As mentioned in Sec. 3.3, the discriminators are trained with R1 regularization $\mathcal{L}_{R1}$~\cite{mescheder2018training} with $\lambda_{R1} = 10$. For training the decoders and the feature dictionaries, we set $\lambda_{n} = 10^{-2}, \lambda_{sdf} = 10^3, \lambda_{rgb} = 10^2, \lambda_{path} = 2, \lambda_{reg} = 10^{-3}$. We select a PCA dimension of $D=16$ for geometry features and $D=8$ for texture features during training. 

For each training iteration, we sample 20480 query points near ground-truth mesh surfaces, and a batch size of $8$ is used. Our decoders and feature dictionaries are trained end-to-end with Adam Optimizer using a learning rate of $0.001$ and first- and second-momentum of $0$ and $0.99$ respectively. The training takes around two days on an RTX 3090 for 8000 epochs for both datasets. 
\begin{figure*}[t]
\centering
\includegraphics[width=\linewidth]{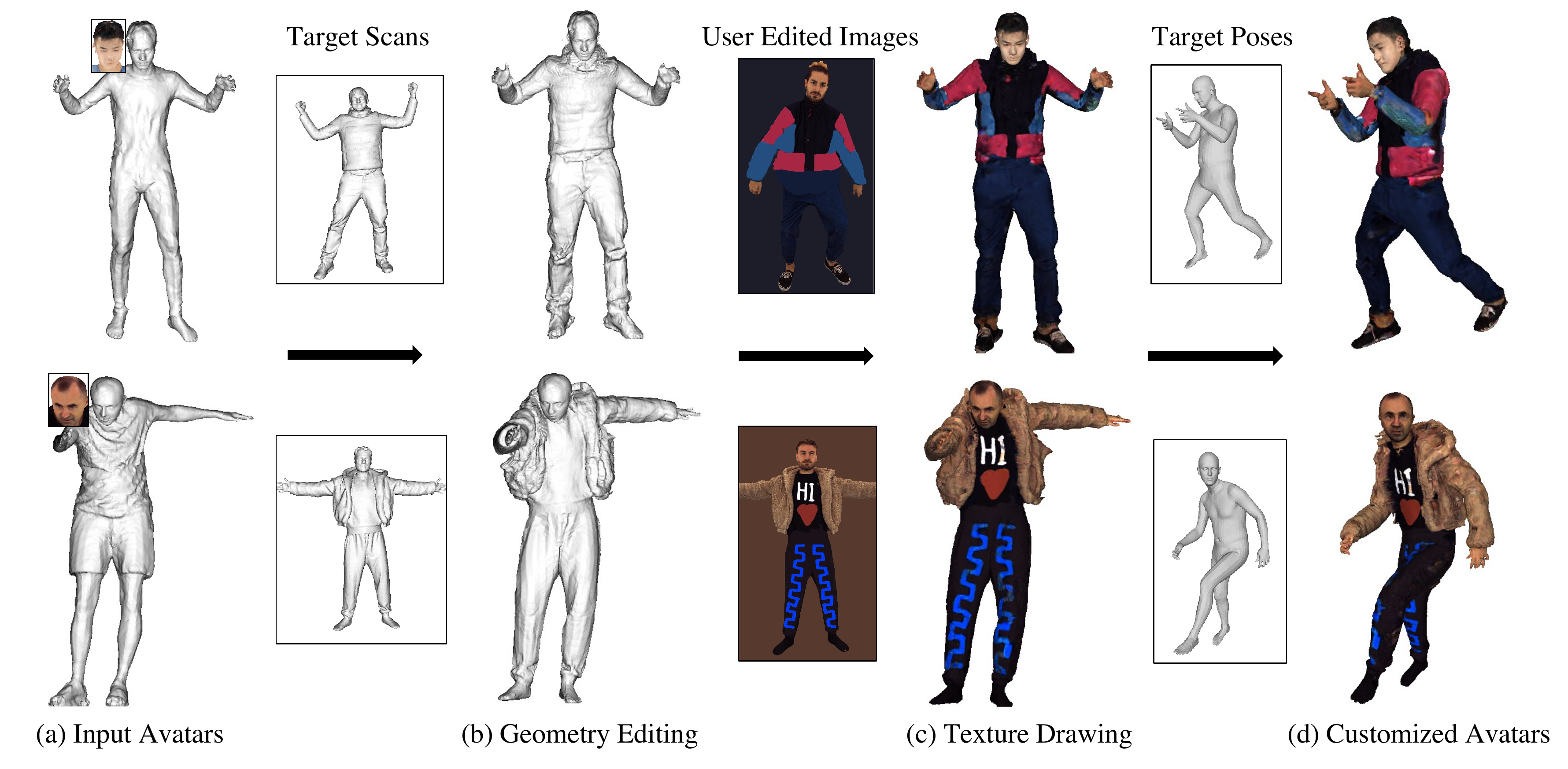}
\caption{\textbf{Avatar customization pipeline}. Two exemplars of avatar creation using our proposed framework. (a) Starting from a random body geometry and texture (\emph{Top}) or training sample saved in our feature dictionary (\emph{Bottom}), we keep human identities (i.e., face textures, geometries) unchanged while gradually (b) editing new clothing geometry and (c) textures onto our avatar by fitting to unseen 3D scans and 2D images. (d) Note that our method can handle challenging loose clothing such as jackets and coats and keep their colors and geometries consistent under various poses. }
\label{fig:main}
\vspace{-1em}
\end{figure*}

\subsection{Inference speed}
Optimizing the geometry converges in 100 iterations ($\sim$40s), fitting textures to a $2$K image takes additional 300 iterations (also $\sim$40s). Our method is currently meant for offline editing; note that our code is unoptimized and can be further accelerated. The resulting meshes are obtained from the predicted SDF values using marching cubes with a resolution of $300^3$ in $\sim$10 seconds.


\section{More Comparisons and Results}

\subsection{Avatar editing}
We present more details and exemplars of customized avatars using our editing workflows in \cref{fig:main}. Starting from a random sample (i.e., $\mathbf{C}_r$ in \cref{fig:main}(a)~\emph{Top}) or from a subject used in training (i.e., $\mathbf{C}_i$ in \cref{fig:main}(a)~\emph{Bottom}), we gradually add new clothing and textures while keeping the identity (i.e., face colors and geometries) and body pose fixed for both cases. First, we invert the target scans into a feature codebook and store it for later use. Note that the body poses of target scans and initial avatars do not need to match. After fitting target scans into codebooks, we change the clothes on the initial avatars by swapping the local features located in the body regions. As shown in \cref{fig:main}(b) our method is able to transfer challenging garments including jackets and preserves local details under different body poses and shapes.
Furthermore, our method allows users to draw personalized logos and letters on garments.  
We keep the body geometry (\cref{fig:main}(b)) unchanged while transferring \emph{only color information} from photos of different subjects and clothing. \cref{fig:main}(c) shows that texture features can be applied to avatars with different poses, shapes, and clothing geometry.
Finally, the resulting avatars enable pose control by changing the SMPL-X parameters without affecting the fitted texture and geometry (\cref{fig:main}(d)).

\subsection{Random sampling}
\begin{figure*}[t]
\centering
\includegraphics[width=0.85\linewidth]{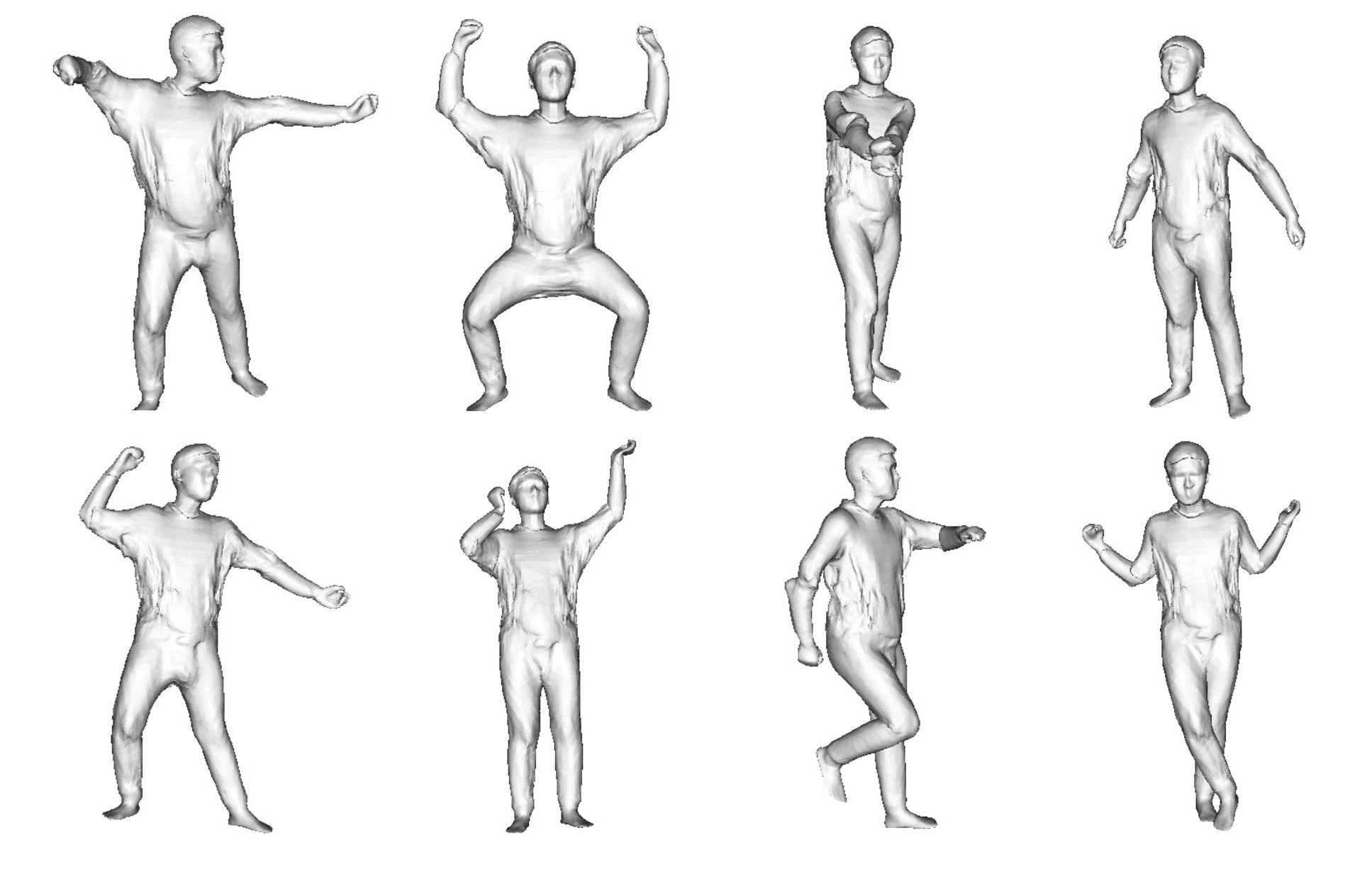}
\caption{\textbf{Random sampled meshes from gDNA~\cite{chen2022gdna}}. The meshes obtained from gDNA do not contain wrinkles and other high-frequency details. }
\label{fig:gdna}
\end{figure*}

\begin{figure*}[h]
\centering
\includegraphics[width=0.85\linewidth]{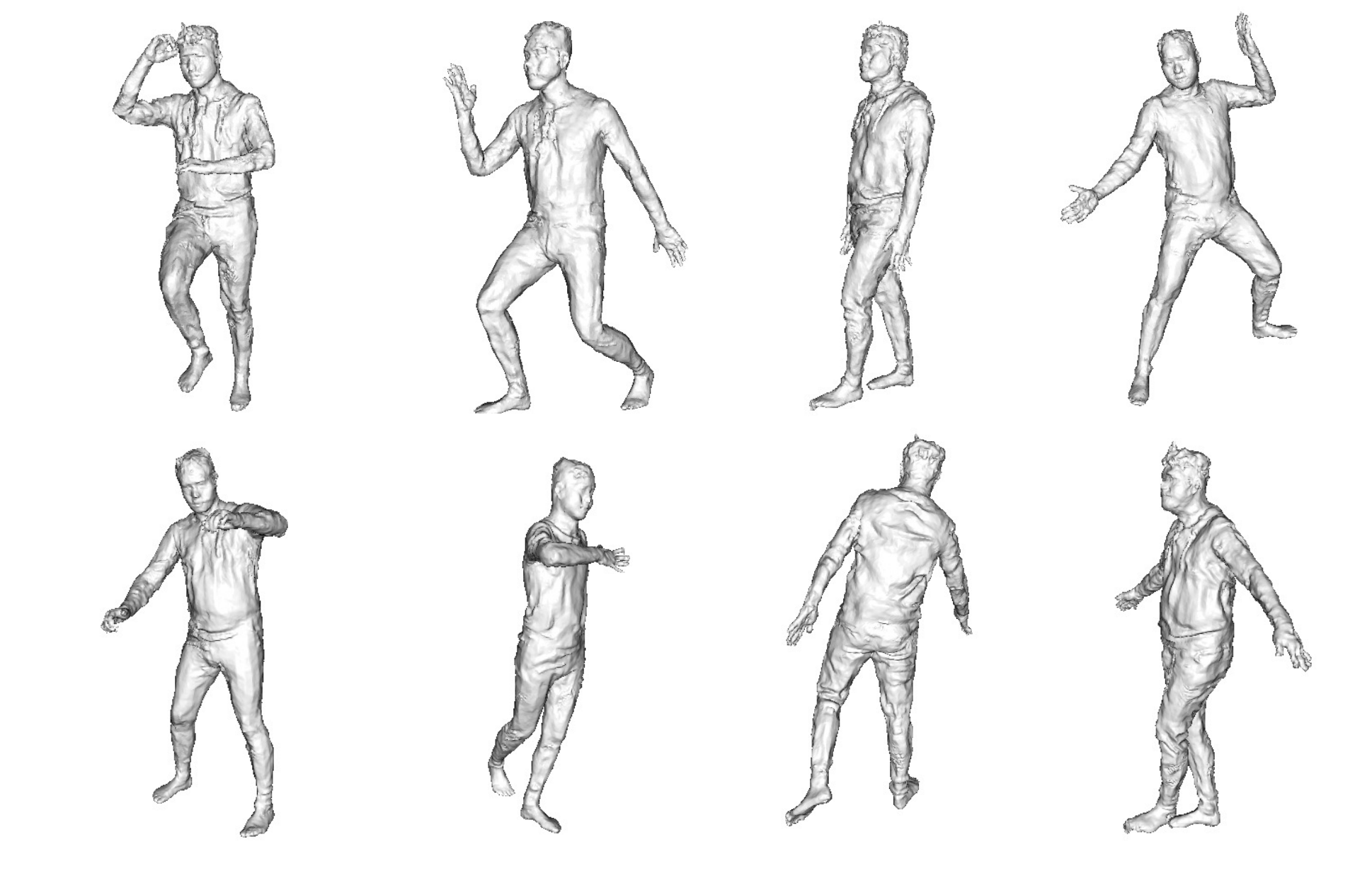}
\caption{\textbf{Random sampled meshes from our model trained on THuman2.0}. Our method can randomly sample meshes that contain more stochastic wrinkles on the clothes and more detailed facial geometries.}
\label{fig:ours}
\end{figure*}

Following the same experiment in Sec. 4.2, we visualize more randomly sampled mesh geometries from gDNA~\cite{chen2022gdna} and our method in \cref{fig:gdna} and \cref{fig:ours}, respectively. As mentioned in the main manuscript, gDNA predicts additional normal maps on the surfaces, thus, their generated meshes do not contain real high-frequency details as shown in \cref{fig:gdna}. Our method, instead, generates more detailed mesh geometries by introducing a 2D adversarial loss during training.

\subsection{Model fitting}
\begin{figure*}[t]
\centering
\includegraphics[width=0.9\linewidth]{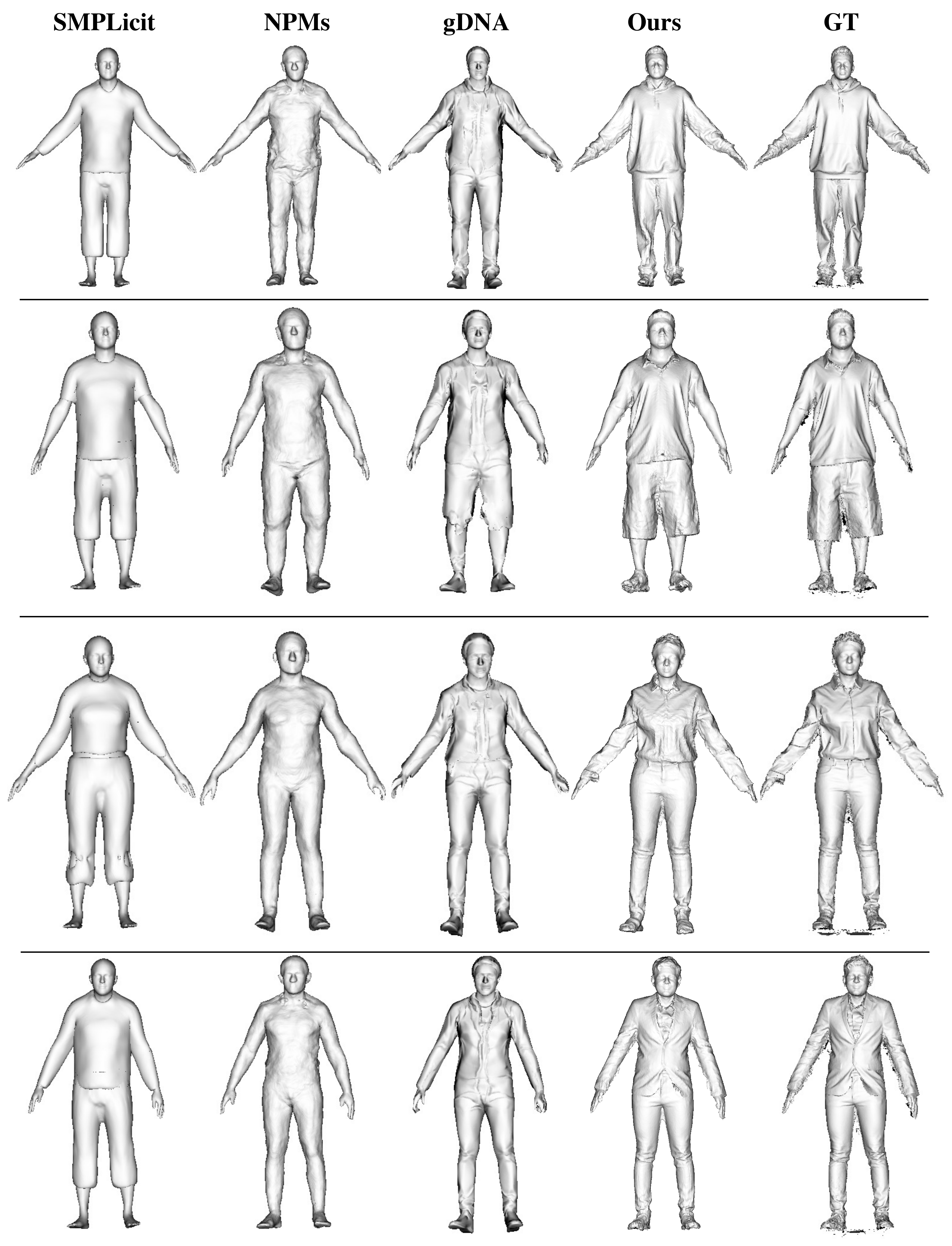}
\caption{\textbf{Qualitative comparison of model fitting on SIZER}. We visualize the fitting results of baselines and our method. Our results are close to the ground truth.}
\label{fig:fitting}
\end{figure*}

Similar to the experiment in Sec. 4.3, we follow the baselines and the selected test subjects introduced in \cite{chen2022gdna} to compare with more human generative baselines, including SMPLicit~\cite{corona2021smplicit}, NPMs~\cite{palafox2021npms}, and gDNA~\cite{chen2022gdna}. We visualize the fitting results of the selected subjects in~\cref{fig:fitting}. In all cases, our results are qualitatively closer to the ground truth compared to other baselines.

\subsection{Reposing}
In \cref{fig:avatar_repose} \emph{Top}, we repose the created avatars using the motion sequence provided in the AIST~\cite{aist-dance-db} dataset. Our representation consistently applies local details to the 3D avatar in different unseen poses.




\begin{figure*}[t]
\centering
\includegraphics[width=0.9\linewidth]{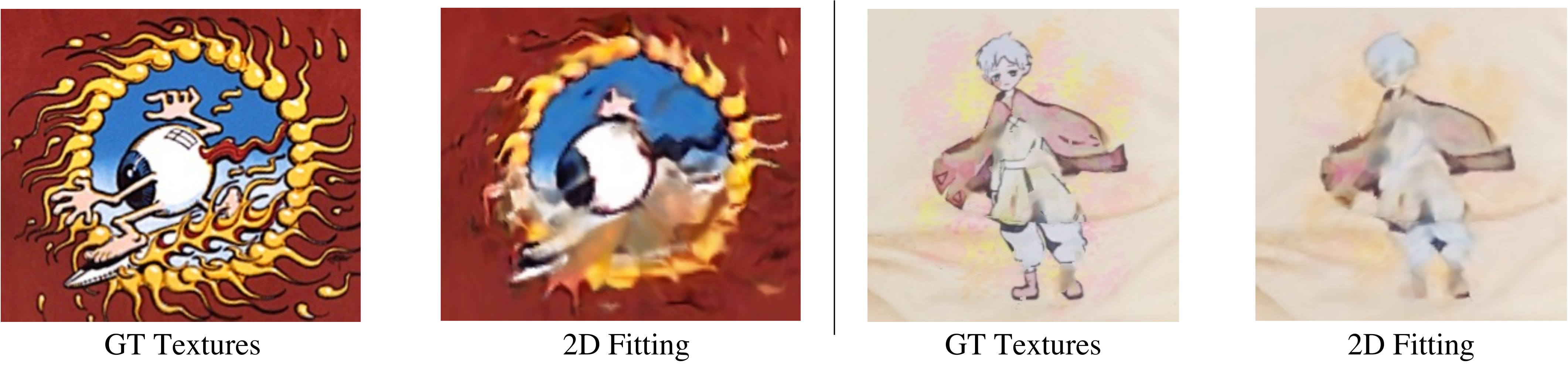}
\caption{\textbf{Imperfect fitting for high-frequency textures}. Due to the limited body mesh resolution, our method cannot fit challenging high-frequency textures very well.}
\label{fig:fail}
\vspace{-1.5em}
\end{figure*}

\section{Discussion}

\subsection{Scalability}
We propose an auto-decoding pipeline that stores learned latent features of $N$ scans in a dictionary. We show that the dictionary size is \emph{not} a major memory bottleneck since each entry occupies 10475*32*2*4 bytes = 2.68MB GPU memory. Even with 1000 scans -- which exceeds the size of the existing datasets -- the memory usage (2.68GB) easily fits onto modern GPUs. In terms of computational time, the PCA operation on a $1000\times320$K dictionary takes 30ms on a single RTX3090. As a reference, one training iteration takes 0.8s. Furthermore, if one wants to apply our method to an even larger dataset (100K), the dictionary and PCA sampling could be replaced by a network that maps a \emph{global feature} to local feature codebooks as is done in~\cite{devries2021unconstrained}.

\subsection{Ethical Concerns}
Data collection and experiments in this work strictly follow the CVPR 2023 Ethics Guidelines. Our data collection procedure has been reviewed and approved by the responsible Institutional Review Board.

Generative modeling and editing of virtual humans are often accompanied by manipulation or fake information. Potential misuse of our creation pipeline to recreate full-body deep fakes for improper applications cannot be fully ruled out even though none of the proposed techniques intend for these purposes. In addition, possible concerns about copyright and privacy might arise since our framework enables the transfer of high-quality local details and facial appearances. Thus, to balance the positive and negative impacts, definite regulations must be established such as creative licenses and personal data protection. By making our code and data publicly available, we hope to raise awareness of future research on detecting fake information on photo-realistic 3D avatars in the Metaverse. 

\subsection{Limitations and future works}
\paragraph{Pose-dependent deformations.}
Our method assumes each human scan is a static observation. Therefore, clothing changes caused by motion and poses cannot be explicitly modeled by local features. However, this is a data- not a model limitation. Once sufficiently large datasets become available, pose-dependent terms can be incorporated into our pipeline similar to~\cite{chen2021snarf}. Hence, one exciting direction is adding pose-dependent terms to the decoders and introducing more synthetic data under various poses to see if the representation and models can extract useful pose-dependent information. 


\paragraph{Challenging textural details.}
As shown in \cref{fig:fail}, our method still suffers from very complex logos. Our implementation uses SMPL-X which limits the modeling of very high-frequency details. Note that our method can be used with \emph{any} mesh template with topological consistency and thus inherently scales to higher resolution representations. Thus, a promising extension of our method would be using a customized human body model that has more vertices on the human body.

\begin{figure*}[t]
\centering
\includegraphics[width=\linewidth]{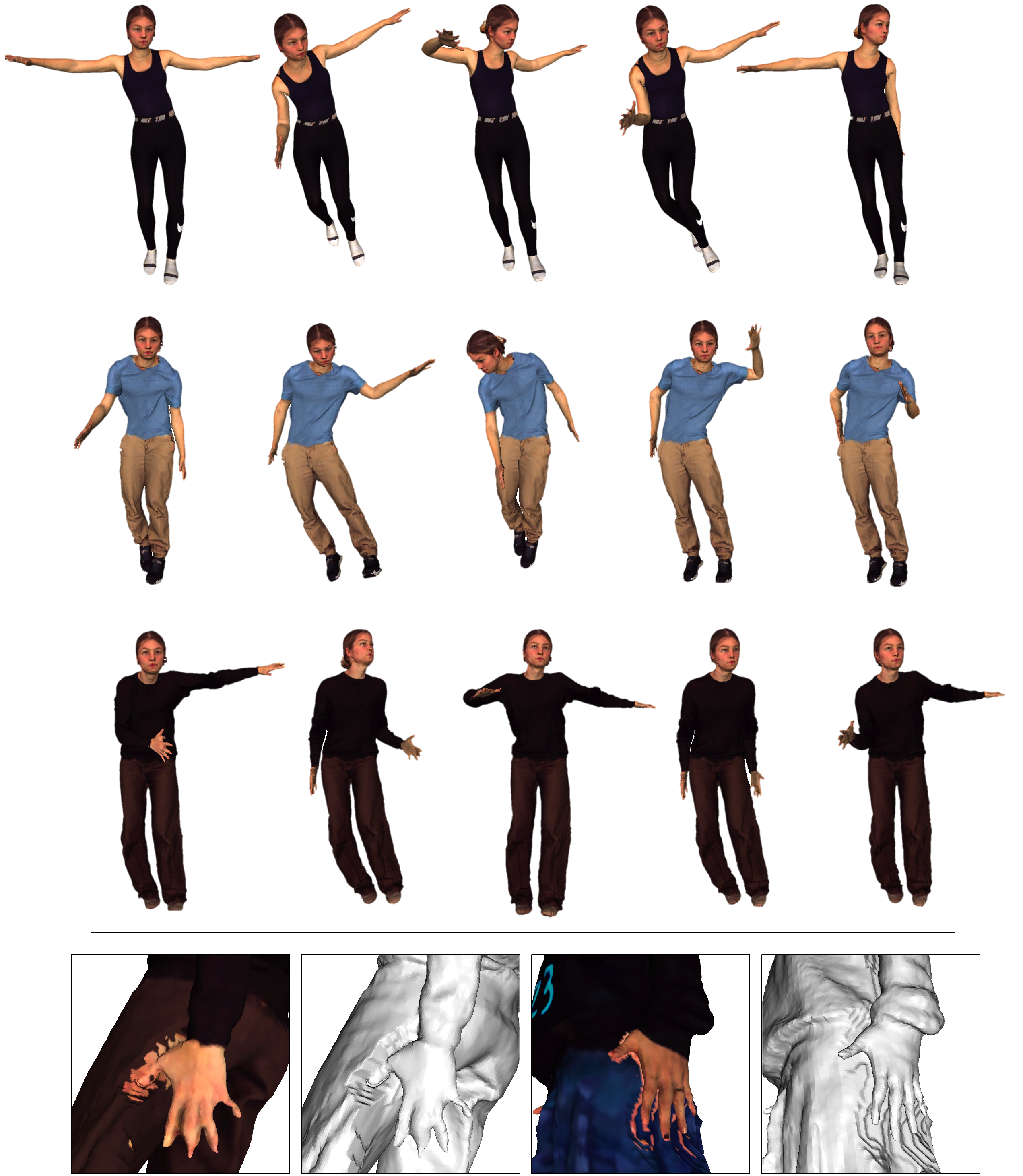}
\caption{\textbf{Avatar reposing}. \emph{Top}: Given edited avatars, our method consistently applies local details stored in the feature codebooks onto the body surfaces in different poses. \emph{Bottom}: Poses with self-contacts might cause artifacts to both texture and geometry.}
\label{fig:avatar_repose}
\vspace{2em}
\end{figure*}

\paragraph{Self-intersection.}
Self-intersections are a well-known issue for learning SDFs since they cause wrong surface samples. To stabilize training, we excluded 3D scans with severe self-contacts in the dataset. As shown in In \cref{fig:avatar_repose} \emph{Bottom}, self-intersections are also an issue in reposing avatars with contact. Therefore, one potential follow-up is to consider both SDF and occupancy and introduce an additional pose-dependent correction for tackling body poses with self-contacts. We also found that in our generalization analysis, training the decoders with more scans and poses reduced artifacts caused by self-intersections. This motivates us to address this issue in a data-driven manner.
\vspace{-1em}
\paragraph{Automatic and interactive editing.} 
Our pipeline still requires users to manually select the vertices of interest for avatar editing which potentially limits the capability of editing complicated attributes such as facial expressions and hairstyles. One possible solution might be automating the current pipeline using a semantic human parsing algorithm as guidance. Moreover, another exciting extension is optimizing the fitting and inference speed and combing an interactive user interface for online avatar creation.


\begin{figure*}[t]
\centering
\includegraphics[width=0.95\linewidth]{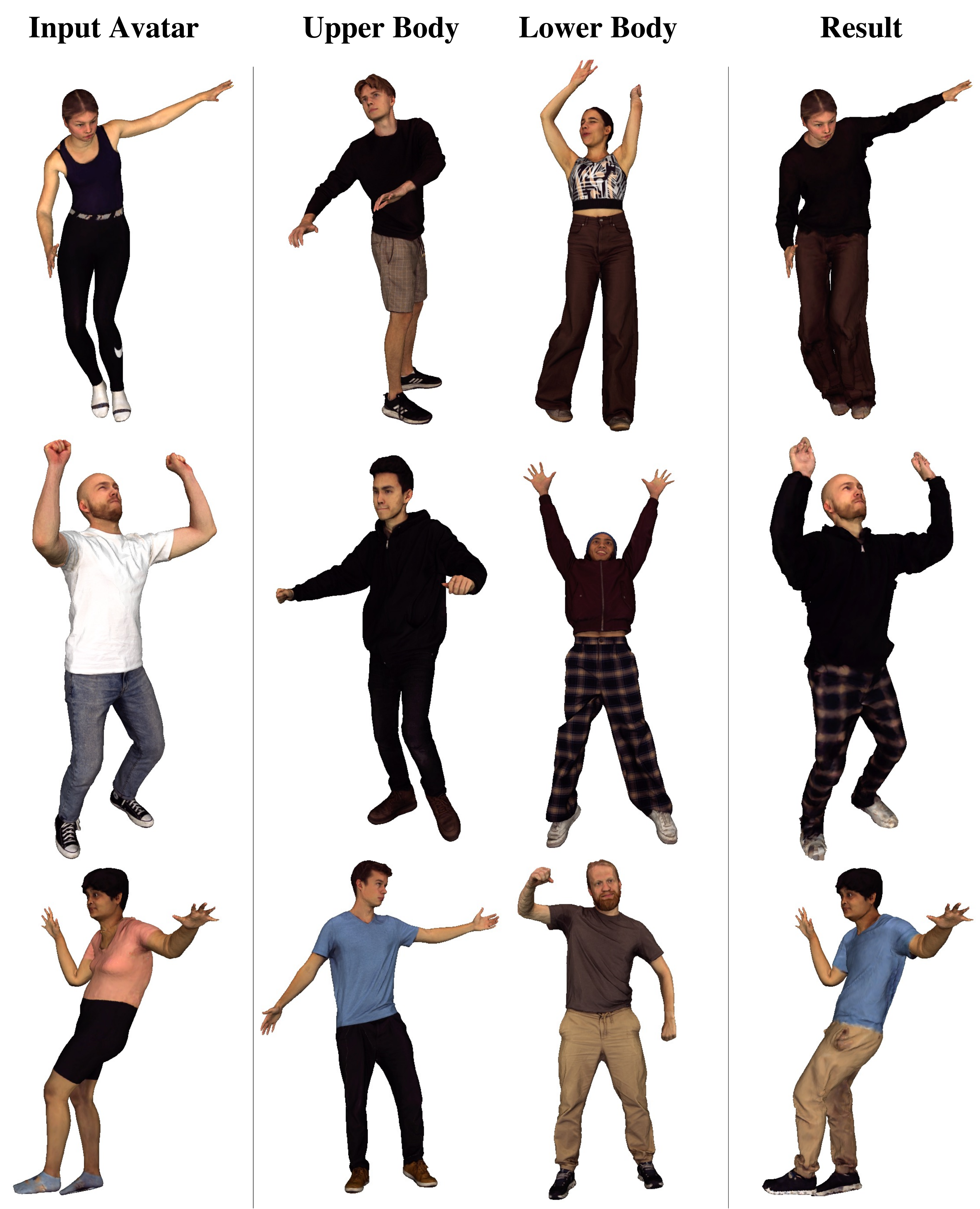}
\caption{\textbf{More cross-subject feature editing results}. We partially transfer local clothing details from the unseen scans (upper and lower body) to the input avatars. The results of the edited avatars are shown in the right column.  }
\label{fig:3D}
\end{figure*}


{\small
\bibliographystyle{ieee_fullname}
\bibliography{egbib2}
}